\documentclass{article}
\usepackage{amsmath}
\usepackage{amsfonts}
\usepackage{amssymb}
\usepackage{textcomp}

\usepackage{microtype}
\usepackage{graphicx}
\usepackage{subcaption}
\usepackage{booktabs} 

\usepackage{hyperref}

\usepackage[accepted]{arxiv-icml2021}

\icmltitlerunning{Transfer of Fully Convolutional Policy-Value Networks Between Games and Game Variants}

\newcommand{\refappendix}[1]{Appendix~\ref{#1}}

\newcommand{\reffigure}[1]{\figurename~\ref{#1}}

\newcommand{\refsubsection}[1]{Subsection~\ref{#1}}
\newcommand{\reftable}[1]{Table~\ref{#1}}

\begin{document}

\twocolumn[
\icmltitle{Transfer of Fully Convolutional Policy-Value Networks\\ Between Games and Game Variants}




\begin{icmlauthorlist} 
\icmlauthor{Dennis J. N. J. Soemers \textsuperscript{*}}{dke}
\icmlauthor{Vegard Mella}{fair}
\icmlauthor{{\'E}ric Piette}{dke}
\icmlauthor{Matthew Stephenson}{dke}
\icmlauthor{Cameron Browne}{dke}
\icmlauthor{Olivier Teytaud}{fair}
\end{icmlauthorlist}

\icmlaffiliation{fair}{Facebook AI Research, Paris, France}
\icmlaffiliation{dke}{Department of Data Science and Knowledge Engineering, Maastricht University, Maastricht, the Netherlands}

\icmlcorrespondingauthor{Dennis Soemers}{dennis.soemers@maastrichtuniversity.nl}

\icmlkeywords{Fully Convolutional, Transfer Learning, General Games}

\vskip 0.3in
]



\printAffiliationsAndNotice{\icmlEqualContribution} 

\newcommand{\otc}[1]{\textcolor{red}{#1}}
\begin{abstract}
In this paper, we  use fully convolutional architectures in AlphaZero-like self-play training setups to facilitate transfer between variants of board games as well as distinct games.
We explore how to transfer trained parameters of these architectures based on shared semantics of channels in the state and action representations of the Ludii general game system. 
We use Ludii's large library of games and game variants for extensive transfer learning evaluations, in zero-shot transfer experiments as well as experiments with additional fine-tuning time.
\end{abstract}

\section{Introduction}
AlphaGo \cite{SilverHuangEtAl16nature} and its successors \cite{Silver2017AlphaGoZero,Silver2018AlphaZero} have inspired a significant amount of research \cite{Anthony2017ExIt,Tian2019ELF,Morandin2019SAI,greatfast,Cazenave2020Polygames,Cazenave2020Mobile} on combinations of self-play, Monte-Carlo Tree Search (MCTS) \cite{Kocsis2006BanditBased,Coulom2007,Browne2012} and Deep Learning \cite{LeCun2015} for automated game-playing. Originally, AlphaGo used distinct value and policy networks, each of which have convolutional layers \cite{LeCun1989CNNs} followed by fully-connected layers. \citet{Silver2017AlphaGoZero} demonstrated that the use of residual blocks \cite{He2016Resnets}, alongside merging the policy and value networks into a single network with two output heads, significantly improved playing strength in AlphaGo Zero. Other modifications to neural network architectures were also explored in subsequent research \cite{greatfast,Cazenave2020Mobile}.

In the majority of prior research, spatial structures present in the state-based inputs for board games are exploited by the inductive bias of convolutional layers, but the policy head -- which has one output for every distinct possible move in a board game -- is preceded by one or more fully-connected layers which do not leverage any spatial semantics. Various architectures that also account for spatial semantics in outputs have been proposed in computer vision literature \cite{unets,fc}, and in the context of games can also handle changes in board size \cite{lin,greatfast,Cazenave2020Polygames}. 

The primary contribution of this paper is an approach for transfer learning between variants of games, as well as distinct games. We use fully convolutional networks with global pooling from the Polygames framework \cite{Cazenave2020Polygames} for their ability to handle changes in spatial dimensions during transfer. We focus on transfer between games implemented in the Ludii general game system \cite{Browne2020Practical,Piette2020Ludii}. Its consistent state and action representations \cite{Piette2020LudiiGameLogicGuide} and game-independent manner of constructing tensor representations for states and actions \cite{Soemers2021DeepLearning} enables the identification of shared semantics between the non-spatial dimensions (channels) of inputs and outputs for different (variants of) games. This facilitates transfer of trained parameters of fully convolutional networks. 
Despite previous publications of benchmarks \cite{openairetro}, transfer learning in games has remained a challenging problem with limited successes. We propose that Ludii's large library of board games can be used as a new benchmark for transfer learning in games, and provide extensive baseline results that include various cases of successful zero-shot transfer and transfer with fine-tuning, for transfer between variants of board games as well as distinct board games.
Our experiments include transfer between domains with differences in board sizes, board shapes, victory conditions, piece types, and other aspects -- many of these have been recognised as important challenges for learning in games \cite{Marcus2018Innateness}.

\section{Background}

This section first provides background information on the implementation of AlphaZero-like \cite{Silver2018AlphaZero} self-play training processes for games in Polygames \cite{Cazenave2020Polygames}, and the architectures of neural networks used for this purpose. Secondly, we briefly describe tensor representations in the Ludii general game system \cite{Browne2020Practical,Piette2020Ludii,Soemers2021DeepLearning}, which we use for all transfer learning experiments. Finally, we discuss background information on transfer learning in games.

\subsection{Learning to Play Games in Polygames} \label{Subsec:PolygamesBackground}

Similar to AlphaZero \cite{Silver2018AlphaZero}, game-playing agents in Polygames \cite{Cazenave2020Polygames} use a combination of MCTS and deep neural networks (DNNs). Experience for training is generated in the form of self-play games between MCTS agents that are guided by the DNN. Given a tensor representation of an input state $s$, the DNN outputs a value estimate $V(s)$ of the value of that state, as well as a discrete probability distribution $\mathbf{P}(s) = \left[ P(s, a_1), P(s, a_2), \dots, P(s, a_n) \right]$ over an action space of $n$ distinct actions. Both of these outputs are used to guide the MCTS-based tree search. The outcomes (typically losses, draws, or wins) of self-play games are used as training targets for the value head (which produces $V(s)$ outputs), and the distribution of visit counts to children of the root node by the tree search process is used as a training target for the policy head (which produces $\mathbf{P}(s)$ outputs).

For board games, input states $s$ are customarily represented as three-dimensional tensors of shape $(C, H, W)$, where $C$ denotes a number of channels, $H$ denotes the height of a $2$D playable area (e.g., a game board), and $W$ denotes the width. The latter two are interpreted as spatial dimensions by convolutional neural networks.
It is typically assumed that the complete action space can be feasibly enumerated in advance, which means that the shape of $\mathbf{P}(s)$ output tensors can be constructed such that every possibly distinct action $a$ has a unique, matching scalar $P(s, a)$ for any possible state $s$ in the policy head. A DNN first produces \textit{logits} $L(s, a)$ for all actions $a$, 
which are transformed into probabilities using a softmax after masking out any actions that are illegal in $s$.

In some general game systems, it can be difficult or impossible to guarantee that there will never be multiple different actions that share a single output in the policy head \cite{Soemers2021DeepLearning} without manually incorporating additional game-specific domain knowledge. We say that distinct actions are \textit{aliased} if they are represented by a single, shared position in the policy head's output tensor. In Polygames, the MCTS visit counts of aliased actions are summed up to produce a single shared training target for the corresponding position in the policy head. In the denominator of the softmax, we only sum over the  distinct logits that correspond to legal actions (i.e., logits for aliased actions are not counted more than once). All aliased actions $a$ receive the same prior probability $P(s, a)$ to bias the tree search -- because the DNN cannot distinguish between them -- but the tree search itself can still distinguish between them.

\subsection{The Ludii General Game System}

Ludii \cite{Browne2020Practical,Piette2020Ludii} is a general game system with over 500 built-in games, many of which support multiple variants with different board sizes, board shapes, rulesets, etc. It automatically constructs suitable object-oriented state and action representations for any game described in its game description language, and these can be converted into tensor representations in a consistent manner without the need for additional game-specific engineering effort \cite{Soemers2021DeepLearning}.
All games in Ludii are modelled as having one or more ``containers'', which can be viewed as areas with spatial semantics (such as boards) that contain relevant elements of game states and positions that are affected by actions. 
This means that all games in Ludii are compatible with fully convolutional architectures.

\subsection{Transfer Learning in Games}

AlphaZero-like training approaches have produced superhuman agents for a variety of board games \cite{Silver2018AlphaZero,Cazenave2020Polygames}, and hence been shown to have fairly general applicability, but models are generally trained from scratch for every distinct game. Transfer learning \cite{Taylor2009TransferRL,Lazaric2012TransferRL,Zhu2020TransferDeepRL} may allow for significant savings in computation costs by transferring trained parameters from a \textit{source domain} (i.e., a game that we train on first) to one or more \textit{target domains} (i.e., variants of the source game or different games altogether).

To the best of our knowledge, research on transfer learning between distinct games has been fairly limited. \citet{Kuhlmann2007GraphBased} investigated the automated discovery of certain types of relations between source and target games, and transferred trained value functions in specific ways for specific relations. \citet{Banerjee2007GeneralGameLearning} proposed to transfer value functions based on game-independent features of a game tree's shape. Both approaches were evaluated, and the former is also restricted to, games defined in the Stanford Game Description Language \cite{love08}.

In this paper, we focus on transferring complete policy-value networks -- with policy as well as value heads -- as they are commonly used in modern game AI literature, using the Ludii general game system to provide easy access to a large and diverse number of games and game variants. 
Crucially, the transfer of trained parameters for a network with a policy head requires the ability to construct a mapping between action spaces of source and target domains, which we propose a method for.
This is in contrast to approaches that only transfer value functions, which also only require the ability to create a mapping between state spaces.

\subsection{Fully Convolutional Architectures}

\begin{figure}
\centering
\includegraphics[width=.95\linewidth]{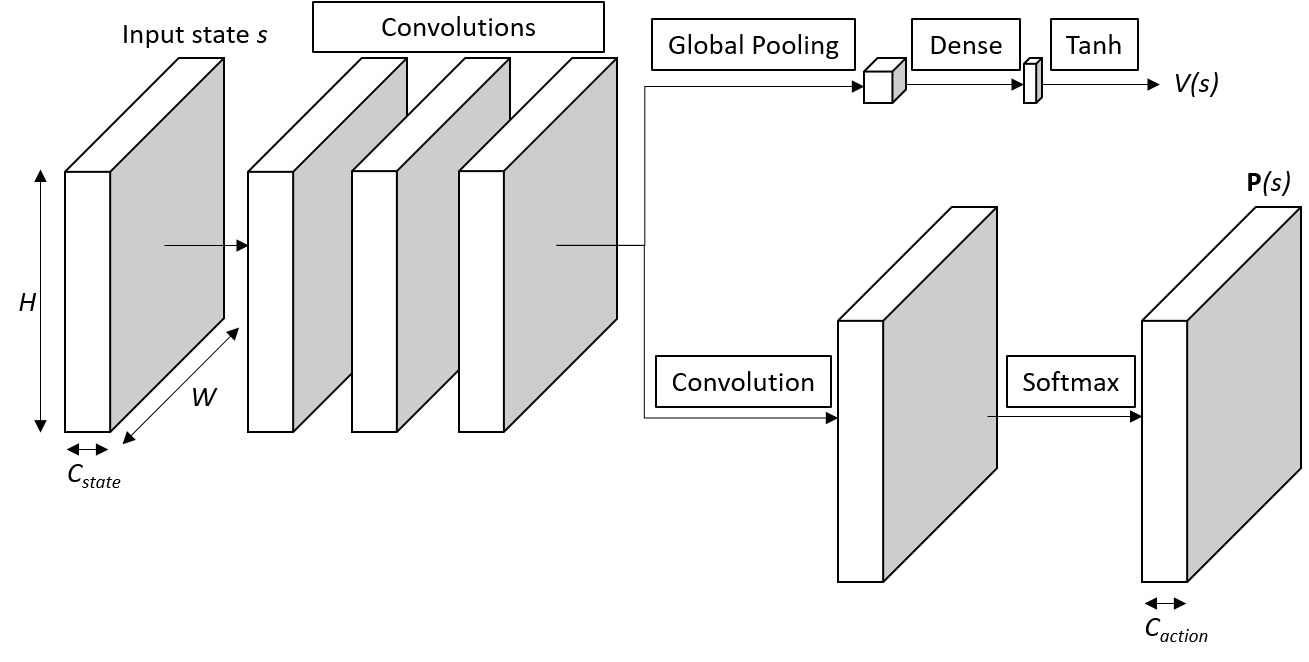}
\caption{Example of a fully convolutional architecture for game playing. Input states $s$ are provided as tensors of shape $(C_{state}, H, W)$. All convolutions construct hidden representations of shape $(C_{hidden}, H, W)$, where $C_{hidden}$ may differ from $C_{state}$. The policy output $\mathbf{P}(s)$ is a tensor of shape $(C_{action}, H, W)$. The value output $V(s)$ is a scalar. Global pooling is used to reduce the dimensionality for the $V(s)$ output in a manner that can transfer to different board sizes \cite{greatfast,Cazenave2020Polygames}.}
\label{Fig:FullyConvArch}
\end{figure}

Following \refsubsection{Subsec:PolygamesBackground}, we assume that game states $s$ are represented as tensors of shape $(C_{state}, H, W)$, and that the action space can be represented by a shape $(C_{action}, H, W)$ -- we may think of actions as being represented by an action channel, and coordinates in the $2$D space of $H$ rows and $W$ columns. A network's policy head therefore also has a shape of $(C_{action}, H, W)$. We focus on $2$-player zero-sum games, which means that a single scalar suffices as output for the value head. Note that different games may have different numbers of channels and different values for $H$ and $W$.

It is common to use architectures that first process input states $s$ using convolutional layers, but at the end use one or more non-convolutional layers (such as fully-connected layers) preceding the outputs \cite{Anthony2017ExIt,Silver2017AlphaGoZero,Silver2018AlphaZero}. This leads to networks that cannot handle changes in the spatial dimensions (i.e., changes in $H$ or $W$), and have no inductive biases that leverage any spatial structure that may be present in the action representation. Fully convolutional architectures with global pooling, such as those implemented in Polygames \cite{Cazenave2020Polygames}, can address both of those concerns. In particular the ability to handle changes in spatial dimensions is crucial for transfer between distinct games or game variants, which is why we restrict our attention to them in the majority of this paper. 

A simplified version of such an architecture is depicted in \reffigure{Fig:FullyConvArch}. Note that some details, such as the use of ReLU activations \cite{Nair2010ReLU}, batch normalization \cite{Ioffe2015BatchNormalization}, and residual connections \cite{He2016Resnets} have been omitted for brevity. Full source code for Polygames \cite{Cazenave2020Polygames}, including its architectures, is available online. \footnote{\url{https://github.com/facebookincubator/Polygames}}

\section{Transferring Parameters Between Games}

The fully convolutional architectures as described in the previous section allow for DNNs trained in a source task $\mathcal{S}$ to be directly used in any target task $\mathcal{T}$ if it only has different values for one or both of the spatial dimensions $H$ and $W$. However, it is also possible that different tasks have different numbers of channels $C_{state}$ or $C_{action}$ for state or action representations (or the same number of channels, but significantly different semantics for those channels). 
This section describes how we identify channels -- for actions as well as states -- that are likely to be semantically ``equivalent'' for \textit{any} pair of games in Ludii's tensor representations \cite{Soemers2021DeepLearning}, and how to transfer trained parameters accordingly. We use a relatively simplistic, binary notion of equivalence; a pair of a channel in $\mathcal{S}$ and a channel in $\mathcal{T}$ will either be considered to have identical semantics, or completely different semantics. Furthermore, we only consider the raw data that is encoded by a channel, and do not account for any differences in game rules in this notion of equivalence. For example, the two channels that encode presence of the two players' pieces in many games such as \textit{Hex}, \textit{Go}, \textit{Tic-Tac-Toe}, etc., are considered to have identical semantics. 
A link to all the source code for identifying these mappings will be provided after double-blind peer review.

\subsection{Mapping State Channels} \label{Subsec:MappingStateChannels}

For the majority of channels in Ludii's state tensor representations \cite{Soemers2021DeepLearning}, semantically equivalent channels are straightforward to identify. For example, a channel that encodes whether player $p$ is the current player, or whether a position was the destination of the previous move in a source domain $\mathcal{S}$, will always have a semantically equivalent channel that encodes identical data and can easily be identified as such in a target domain $\mathcal{T}$. These cases are listed in detail in Appendix A of the supplementary material.
The first non-trivial case is that of binary channels which encode, for every pair of spatial coordinates, whether or not the corresponding position exists in a container represented by that channel. For example, the game of \textit{Shogi} in Ludii has three separate containers; a large container for the game board, and two smaller containers representing player ``hands'', which hold captured pieces. These different containers are assigned different sections of the $2$D space, and every container has a binary channel that tells the DNN which positions exist in which containers. In all the built-in games available in Ludii, containers are ordered in a ``semantically consistent'' manner; the first container is always the main game board, always followed by player hands (if there are any), etc. Therefore, we use the straightforward approach of mapping these container-based channels simply in the order in which they appear. 
It may be possible to remove the reliance on such domain knowledge with techniques that automatically analyse the rules of a game in more detail \cite{Kuhlmann2007GraphBased,BouAmmar2013AutomatedTransfer,BouAmmar2014MDPSimilarity}, but is outside the scope of this work.

The second case that warrants additional explanation is that of binary channels that encode the presence of pieces; for every distinct piece type that is defined in a game, Ludii creates a binary channel that indicates for every possible position whether or not a piece of that type is placed in that position. Many games (such as \textit{Hex}, \textit{Go}, \textit{Tic-Tac-Toe}, etc.) only have a single piece type per player, and for these we could easily decide that a channel indicating presence of pieces of Player $p$ in one game is semantically equivalent to a channel indicating presence of pieces of the same player in another game. 
For cases with more than a single piece type per player, we partially rely on an unenforced convention that pieces in built-in games of Ludii tend to be named consistently across closely-related games (e.g., similar piece type names are used in many variants of \textit{Chess}).
If we find an exact name match for piece types between $\mathcal{S}$ and $\mathcal{T}$, we treat the corresponding channels as semantically equivalent. Otherwise, for any piece type $j$ in $\mathcal{T}$, we loop over all piece types in $\mathcal{S}$, and compute the Zhang-Shasha tree edit distances \cite{ZhangShasha1989TreeEditDistance} between the trees of ``ludemes'' that describe the rules for the piece types in Ludii's game description language \cite{Piette2020Ludii}. 

\subsection{Transferring State Channel Parameters} \label{Subsec:TransferringStateParams}

For the purpose of determining how to transfer parameters that were trained in $\mathcal{S}$ to a network that can play $\mathcal{T}$, we make the assumption that $\mathcal{S}$ and $\mathcal{T}$ \textit{are exactly the same game, but with different state tensor representations}; some state channels may have been added, removed, or shuffled around. We make this assumption because it enables us to build a more rigorous notion of what it means to ``correctly'' transfer parameters without accounting for differences in rules, optimal strategies, or value functions. In practice, $\mathcal{S}$ and $\mathcal{T}$ can end up being different games, and we intuitively still expect this transfer to be potentially beneficial if the games are sufficiently similar, but the notion of ``correct'' transfer cannot otherwise be made concrete without detailed domain knowledge of the specific games involved.

Let $s$ denote the tensor representation, of shape $(C_{state}^{\mathcal{S}}, H, W)$, for any arbitrary state in $\mathcal{S}$. Let $s'$ denote the tensor representation, of shape $(C_{state}^{\mathcal{T}}, H, W)$, for the same state represented in $\mathcal{T}$ -- this must exist by the assumption that $\mathcal{S}$ and $\mathcal{T}$ are the same game, modelled in different ways. Let $h_1^{\mathcal{S}}(s)$ denote the hidden representation obtained by the application of the first convolutional operation on $s$, in a network trained on $\mathcal{S}$. For brevity we focus on the case used throughout all our experiments, but most if not all of these assumptions can likely be relaxed; a \texttt{nn.Conv2d} layer as implemented in PyTorch \cite{Paszke2019PyTorch}, with $3$$\times$$3$ filters, a stride and dilation of $1$, and a padding of $1$ (such that the spatial dimensions do not change). Let $\Theta^{\mathcal{S}}$ denote this layer's tensor of parameters trained in $\mathcal{S}$, of shape $(k_{out}, C_{state}^{\mathcal{S}}, 3, 3)$, and $B^{\mathcal{S}}$ -- of shape $(k_{out})$ -- the corresponding bias. This leads to $h_1^{\mathcal{S}}$ having a shape of $(k_{out}, H, W)$. Similarly, let $h_1^{\mathcal{T}}(s')$ denote the first hidden representation in the network after transfer, for the matching state $s'$ in the new target domain's representation, with weight and bias tensors $\Theta^{\mathcal{T}}$ and $B^{\mathcal{T}}$. 

Under the assumption of source and target games being identical, we could obtain correct transfer by ensuring that $h_1^{\mathcal{S}}(s) = h_1^{\mathcal{T}}(s')$. Achieving this would mean that the first convolutional layer would handle any changes in the state representation, and the remainder of the network could be transferred in its entirety and behave as it learned to do in the source domain. $B^{\mathcal{T}}$ is simply initialised by copying $B^{\mathcal{S}}$. Let $i \cong j$ denote that the $i^{th}$ channel in a source domain $\mathcal{S}$ has been determined (as described in \refsubsection{Subsec:MappingStateChannels}) to be semantically equivalent to the $j^{th}$ channel in a target domain $\mathcal{T}$. Channels on the left-hand side are always source domain channels, and channels on the right-hand side are always target domain channels. For any channel $j$ in $\mathcal{T}$, if there exists a channel $i$ in $\mathcal{S}$ such that $i \cong j$, we initialise $\Theta^{\mathcal{T}}(k, j, \cdot, \cdot) := \Theta^{\mathcal{S}}(k, i, \cdot, \cdot)$ for all $k$. If there is no such channel $i$, we initialise these parameters using the default approach for initialising untrained parameters (or initialise them to $0$ for zero-shot evaluations, where these parameters do not need to remain trainable through backpropagation).

If $\mathcal{T}$ contains channels $j$ such that there are no equivalent channels $i$ in $\mathcal{S}$, i.e. $\{ i \mid i \cong j \} = \varnothing$, we have no transfer to the $\Theta^{\mathcal{T}}(k, j, \cdot, \cdot)$ parameters. This can be the case if $\mathcal{T}$ involves new data for which there was no equivalent in the representation of $\mathcal{S}$, which means that there was also no opportunity to learn about this data in $\mathcal{S}$.

If $\mathcal{S}$ contained channels $i$ such that there are no equivalent channels $j$ in $\mathcal{T}$, i.e. $\nexists j \left( i \cong j \right)$, we have no transfer from the $\Theta^{\mathcal{S}}(k, i, \cdot, \cdot)$ parameters. This can be the case if $\mathcal{S}$ involved data that is no longer relevant or accessible in $\mathcal{T}$. Not using them for transfer is equivalent to pretending that these channels still are present, but always filled with $0$ values.

If $\mathcal{S}$ contained a channel $i$ such that there are multiple equivalent channels $j$ in $\mathcal{T}$, i.e. $\left| \{ j \mid i \cong j \} \right| > 1$, we copy a single set of parameters multiple times. This can be the case if $\mathcal{T}$ uses multiple channels to encode data that was encoded by just a single channel in $\mathcal{S}$ (possibly with a loss of information). In the case of Ludii's tensor representations, this only happens when transferring channels representing the presence of piece types from a game $\mathcal{S}$ with fewer types of pieces, to a game $\mathcal{T}$ with more distinct piece types. In the majority of games in Ludii, such channels are ``mutually exclusive'' in the sense that if a position contains a $1$ entry in one of these channels, all other channels in the set are guaranteed to have a $0$ entry in the same position. This means that copying the same parameters multiple times can still be viewed as ``correct''; for any given position in a state, they are guaranteed to be multiplied by a non-zero value at most once. The only exceptions are games $\mathcal{T}$ that allow for multiple pieces of distinct types to be stacked on top of each other in a single position, but these games are rare and not included in any of the experiments described in this paper.

If $\mathcal{T}$ contains a single channel $j$ such that there are multiple equivalent channels $i$ in $\mathcal{S}$, i.e. $\left| \{ i \mid i \cong j \} \right| > 1$, there is no clear way to correctly transfer parameters without incorporating additional domain knowledge on how the single target channel summarises -- likely with a loss of information -- multiple source channels. This case never occurs when mapping channels as described in \refsubsection{Subsec:MappingStateChannels}.

\subsection{Mapping Action Channels}

Channels in Ludii's action tensor representations \cite{Soemers2021DeepLearning} have three broad categories of channels; a channel for pass moves, a channel for swap moves, and one or more channels for all other moves. Channels for pass or swap moves in one domain can easily be classified as being semantically equivalent only to channels for the same type of moves in another domain.

We refer to games where, in Ludii's internal move representation \cite{Piette2020LudiiGameLogicGuide}, some moves have separate ``from'' and ``to'' (or source and destination) positions as \textit{movement games} (e.g. \textit{Amazons}, \textit{Chess}, \textit{Shogi}, etc.), and games where all moves only have a ``to'' position as \textit{placement games} (e.g. \textit{Go}, \textit{Hex}, \textit{Tic-Tac-Toe}, etc.). In placement games, there is only one more channel to encode all moves that are not pass or swap moves. In movement games, there are $49$ additional channels, which can distinguish moves based on any differences in $x$ and $y$ coordinates between ``from'' and ``to'' positions in $\{\leq -3, -2, -1, 0, 1, 2, \geq 3 \}$.

If both $\mathcal{S}$ and $\mathcal{T}$ are placement games, or if both are movement games, we can trivially obtain one-to-one mappings between all move channels. If $\mathcal{S}$ is a movement game, but $\mathcal{T}$ is a placement game, we only treat the source channel that encodes moves with equal ``from'' and ``to'' positions as semantically equivalent (in practice, this channel remains unused in the vast majority of movement games, which means that we effectively get no meaningful transfer for moves due to the large discrepancy in movement mechanisms). If $\mathcal{S}$ is a placement game, and $\mathcal{T}$ is a movement game, we treat the sole movement channel from $\mathcal{S}$ as being semantically equivalent to \textit{all} the movement channels in $\mathcal{T}$.

\subsection{Transferring Action Channel Parameters}

Similar to \refsubsection{Subsec:TransferringStateParams}, we make the assumption that source and target games $\mathcal{S}$ and $\mathcal{T}$ are identical games, but with \textit{different action tensor representations}, such that we can define a clear notion of correctness for transfer. Let $s$ and $a$ denote any arbitrary state and action in $\mathcal{S}$, such that $a$ is legal in $s$, and let $s'$ and $a'$ denote the corresponding representations in $\mathcal{T}$. We assume that the state representations have been made equivalent through transfer of parameters for the first convolutional layer, as described in \refsubsection{Subsec:TransferringStateParams}.
Let $h_n(s)$ be the hidden representation that, in a fully convolutional architecture trained in $\mathcal{S}$, is transformed into a tensor $L(h_n(s))^{\mathcal{S}}$ of shape $(C_{action}^{\mathcal{S}}, H, W)$ of logits. Similarly, let $L(h_n(s'))^{\mathcal{T}}$ of shape $(C_{action}^{\mathcal{T}}, H, W)$ denote such a tensor of logits in the target domain. By assumption, we have that $h_n(s) = h_n(s')$.

If the action representations in $\mathcal{S}$ and $\mathcal{T}$ are equally powerful in their ability to distinguish actions, we can define a notion of correct transfer of parameters by requiring the transfer to ensure that $L(h_n(s))^{\mathcal{S}} = L(h_n(s'))^{\mathcal{T}}$ after accounting for any shuffling of channel indices. If we have one-to-one mappings for all action channels, this can be easily achieved by copying parameters of the final convolutional operation in a similar way as for state channels (see \refsubsection{Subsec:TransferringStateParams}).

If the action representation of $\mathcal{T}$ can distinguish actions from each other that cannot be distinguished in $\mathcal{S}$, we have a reduction in \textit{move aliasing} (see \refsubsection{Subsec:PolygamesBackground}). This happens when transferring from placement games to movement games. Since these actions were treated as identical when training in $\mathcal{S}$, it is sensible to continue treating them as identical and give them equal probabilities in $\mathcal{T}$. This is achieved by mapping a single source channel to multiple target channels, and copying trained parameters accordingly. 

If the action representation of $\mathcal{S}$ could distinguish actions from each other that can no longer be distinguished in $\mathcal{T}$. This happens when transferring from movement games to placement games. As described in the previous subsection, we handle this case conservatively by only allowing transfer from a single source channel -- the one that is arguably the ``most similar'' -- and discarding all other parameters.

\section{Experiments} \label{Sec:Experiments}

\begin{figure*}
\centering
\includegraphics[width=.95\textwidth]{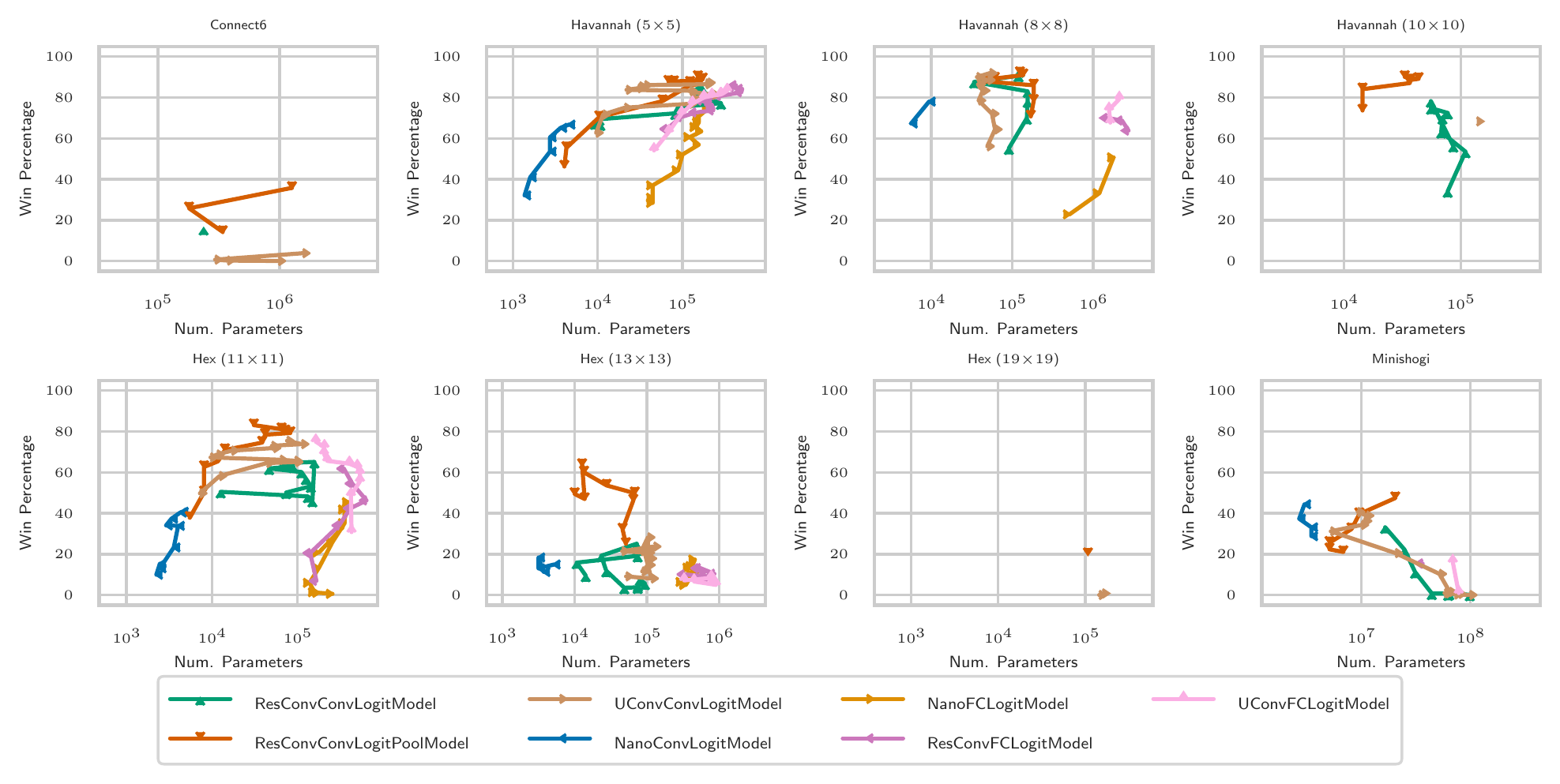}
\caption{Win percentages of a trained MCTS with 40 iterations/move vs. UCT with 800 iterations/move for a variety of architectures. \texttt{Nano}: shallow architectures. \texttt{ConvConv}: deep fully convolutional. \texttt{ConvFC}:  fully connected layers after convolutional ones. \texttt{Pool}: adding global pooling; \texttt{U}: adding U-connections \cite{unets}. Deep is better than shallow, U-nets are slightly better than their classical counterparts, and deep nets are greatly improved by (i) using fully convolutional policy heads (ii) using global pooling. 
}
\label{Fig:ResultsFullyConvArchitectures}
\end{figure*}

This section discusses experiments used to evaluate the performance of fully convolutional architectures, as well as several transfer learning experiments between variants of games and between distinct games. We used the training code from Polygames \cite{Cazenave2020Polygames}. For transfer learning experiments, we used games as implemented in Ludii v1.1.6 \cite{Browne2020Practical}. Appendix B of the supplementary material provides details on hyperparameters.

\subsection{Evaluation of Fully Convolutional Architectures}

We selected a variety of board games as implemented in Polygames \cite{Cazenave2020Polygames}, and trained networks of various sizes and architectures, using 24 hours on 8 GPUs and 80 CPU cores per model. Models of various sizes -- measured by the number of trainable parameters -- have been constructed by randomly drawing choices for hyperparameters such as the number of layers, blocks, and channels for hidden layers. After training, we evaluated the performance of every model by recording the win percentage of an MCTS agent using 40 iterations per move with the model, versus a standard untrained UCT \cite{Browne2012} agent with 800 iterations per move. These win percentages are depicted in \reffigure{Fig:ResultsFullyConvArchitectures}. In the majority of 
cases,
\texttt{ResConvConvLogitPoolModel} -- a fully convolutional model with global pooling -- is among the strongest architectures. Fully convolutional models generally outperform ones with dense layers, and models with global pooling generally outperform those without global pooling. This suggests that using such architectures can be beneficial in and of itself, and their use to facilitate transfer learning does not lead to a sacrifice in baseline performance.

\subsection{Evaluation of Transfer Learning}

All transfer learning experiments discussed below used the \texttt{ResConvConvLogitPoolModelV2} architecture from Polygames \cite{Cazenave2020Polygames}. All models were trained for 20 hours on 8 GPUs and 80 CPU cores, using 1 server for training and 7 clients for the generation of self-play games.

\subsubsection{Transfer Between Game Variants}

We selected a set of nine different board games, as implemented in Ludii, and for each of them consider a few different variants. The smallest number of variants for a single game is $2$, and the largest number of variants for a single game is $6$. In most cases, the different variants are simply different board sizes. For example, we consider \textit{Gomoku} played on $9$$\times$$9$, $13$$\times$$13$, $15$$\times$$15$, and $19$$\times$$19$ boards as four different variants of \textit{Gomoku}. We also include some cases where board shapes change (i.e., \textit{Breakthrough} played on square boards as well as hexagonal boards), ``small'' changes in rules (i.e., \textit{Broken Line} played with a goal line length of $3$, $4$, $5$, or $6$), and ``large'' changes in rules (i.e., \textit{Hex} with the standard win condition and \textit{Mis{\`e}re Hex} with an inverted win condition). Details on all the games and game variants used are provided in Appendix C of the supplementary material.

We trained a separate model for every variant of each of these games, and within each game, transferred models from all variants to all other variants. We evaluate zero-shot transfer performance for a source domain $\mathcal{S}$ and target domain $\mathcal{T}$ by reporting the win percentage of the model trained in $\mathcal{S}$ against the model that was trained in $\mathcal{T}$, over $300$ evaluation games per $(\mathcal{S}, \mathcal{T})$ tuple running in $\mathcal{T}$.

\begin{figure}
    \centering
    \includegraphics[width=.76\linewidth]{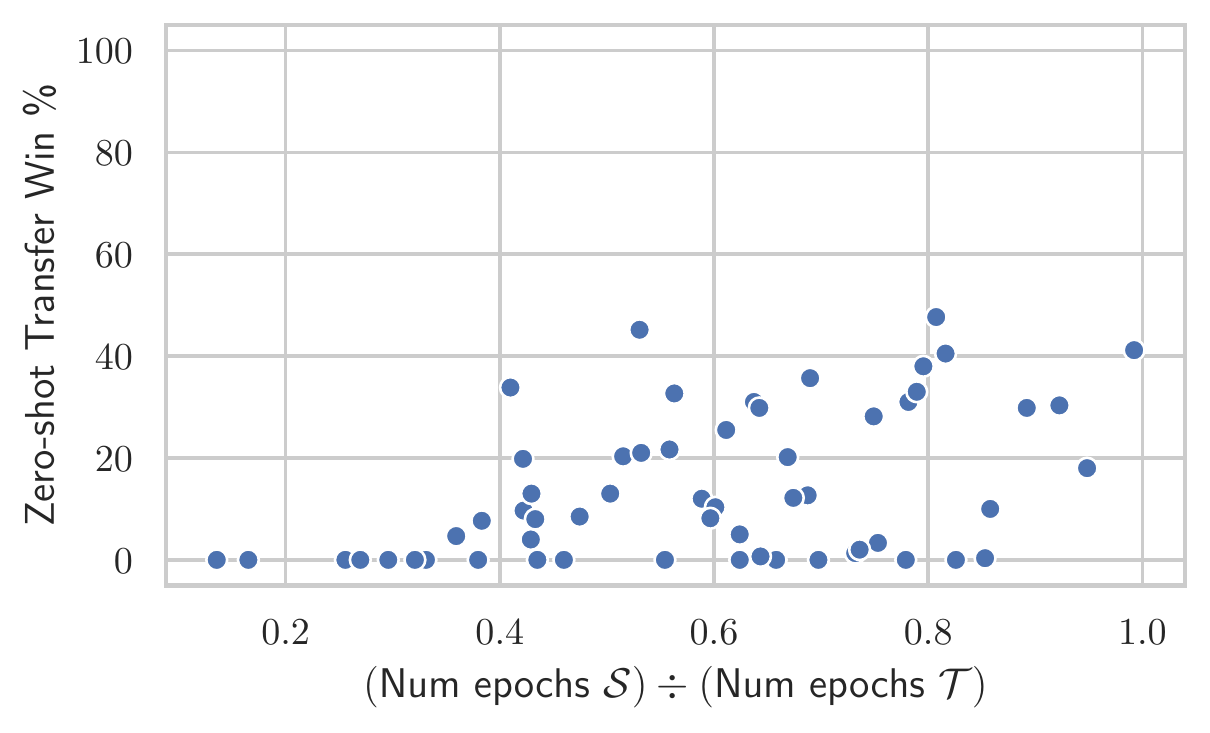}
    \caption{Zero-shot transfer from $\mathcal{S}$ with larger board sizes to $\mathcal{T}$ with smaller board sizes, for several board games and board sizes. 
    }
    \label{Fig:ZeroShotBoardSizeDecreases}
\end{figure}

\begin{figure}
    \centering
    \includegraphics[width=.76\linewidth]{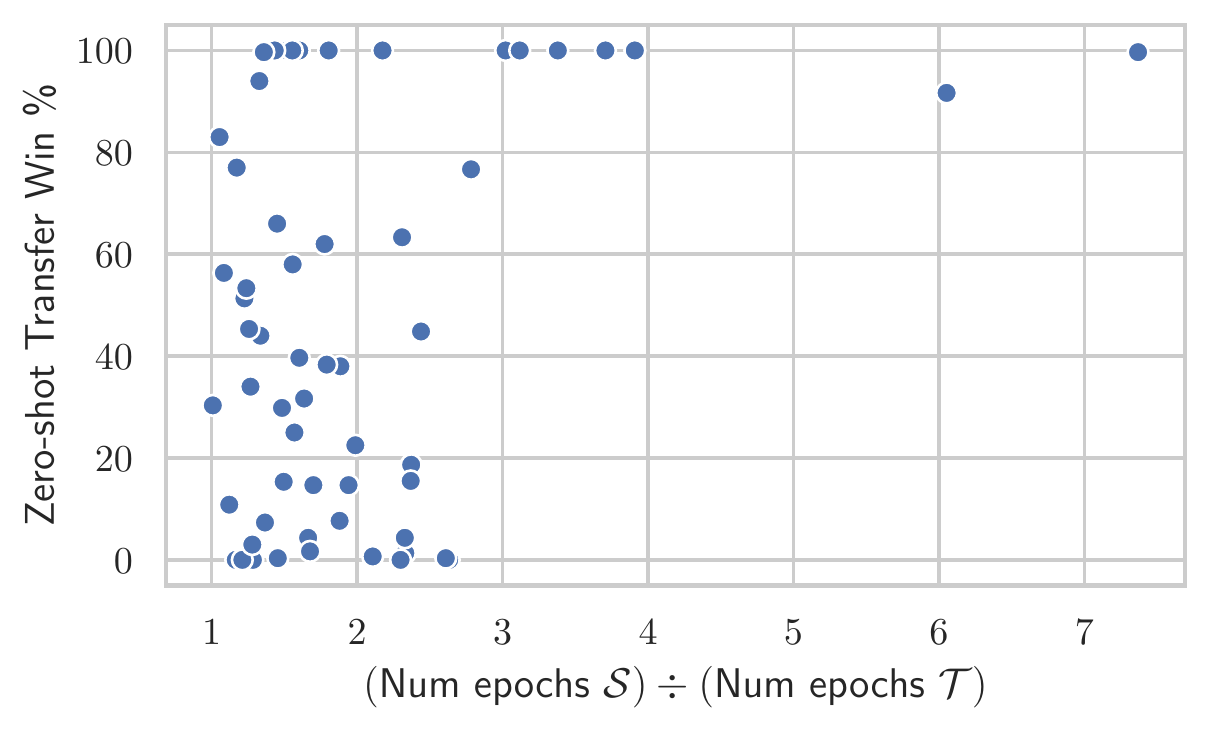}
    \caption{Zero-shot transfer from $\mathcal{S}$ with smaller board sizes to $\mathcal{T}$ with larger board sizes, for several board games and board sizes.}
    \label{Fig:ZeroShotBoardSizeIncreases}
\end{figure}

\reffigure{Fig:ZeroShotBoardSizeDecreases} and \reffigure{Fig:ZeroShotBoardSizeIncreases} depict scatterplots of all these zero-shot transfer evaluations for the cases where $\mathcal{S}$ has a larger board size than $\mathcal{T}$, and where $\mathcal{S}$ has a smaller board size than $\mathcal{T}$, respectively. The $y$-axis represents win percentages of the transferred model against the baseline model, and the $x$-axis represents the ratio of the number of training epochs of the source model to the number of training epochs of the target model. Models trained on larger board sizes tend to have a lower number of training epochs for three reasons; the neural network passes are more expensive, the game logic in Ludii is more expensive, and episodes often tend to last for a higher number of turns when played on larger boards. Hence, all data points in \reffigure{Fig:ZeroShotBoardSizeDecreases} have a ratio $\leq 1.0$, and all data points in \reffigure{Fig:ZeroShotBoardSizeIncreases} have a ratio $\geq 1.0$. 

When transferring a model that was trained on a large board to a small board (\reffigure{Fig:ZeroShotBoardSizeDecreases}), zero-shot win percentages tend to be below $50\%$, but frequently still above $0\%$. This suggests that training on a larger board than the one we intend to play on does not outperform simply training on the correct board directly, but it often still produces a capable model that can win a non-trivial number of games. When transferring a model that was trained on a small board to a large board (\reffigure{Fig:ZeroShotBoardSizeIncreases}), we also frequently obtain win percentages above $50\%$, even reaching up to $100\%$, against models that were trained directly on the board used for evaluation.

Zero-shot transfer between variants with larger differences, such as modified board shapes or changes in win conditions, only leads to win percentages significantly above $0\%$ in a few cases. These results, as well as more detailed tables,
are presented in Appendix D of the supplementary material.

\begin{figure*}
\centering
\begin{subfigure}{.34\textwidth}
  \centering
  \includegraphics[width=\linewidth]{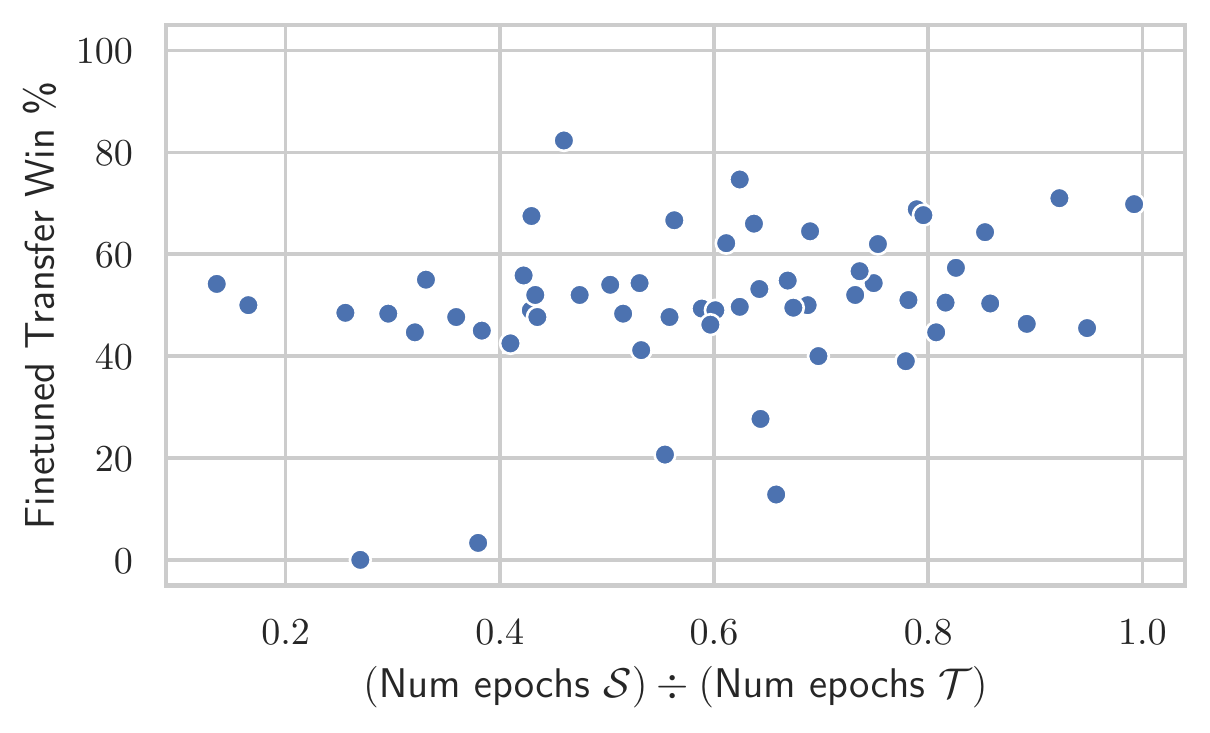}
  \caption{$\mathcal{S}$ with larger board sizes than $\mathcal{T}$.}
  \label{Fig:ResultsFinetuningBoardSizeDecreases}
\end{subfigure}
\begin{subfigure}{.34\textwidth}
  \centering
  \includegraphics[width=\linewidth]{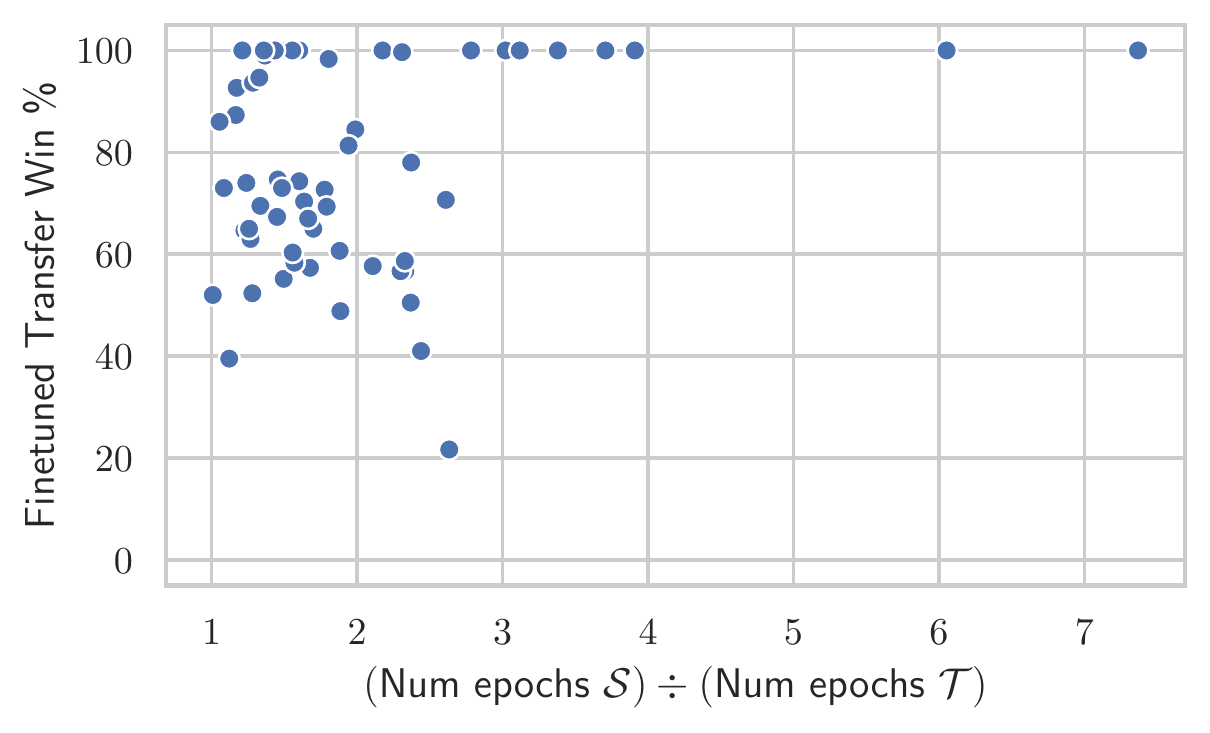}
  \caption{$\mathcal{S}$ with smaller board sizes than $\mathcal{T}$.}
  \label{Fig:ResultsFinetuningBoardSizeIncreases}
\end{subfigure}
\begin{subfigure}{.34\textwidth}
  \centering
  \includegraphics[width=\linewidth]{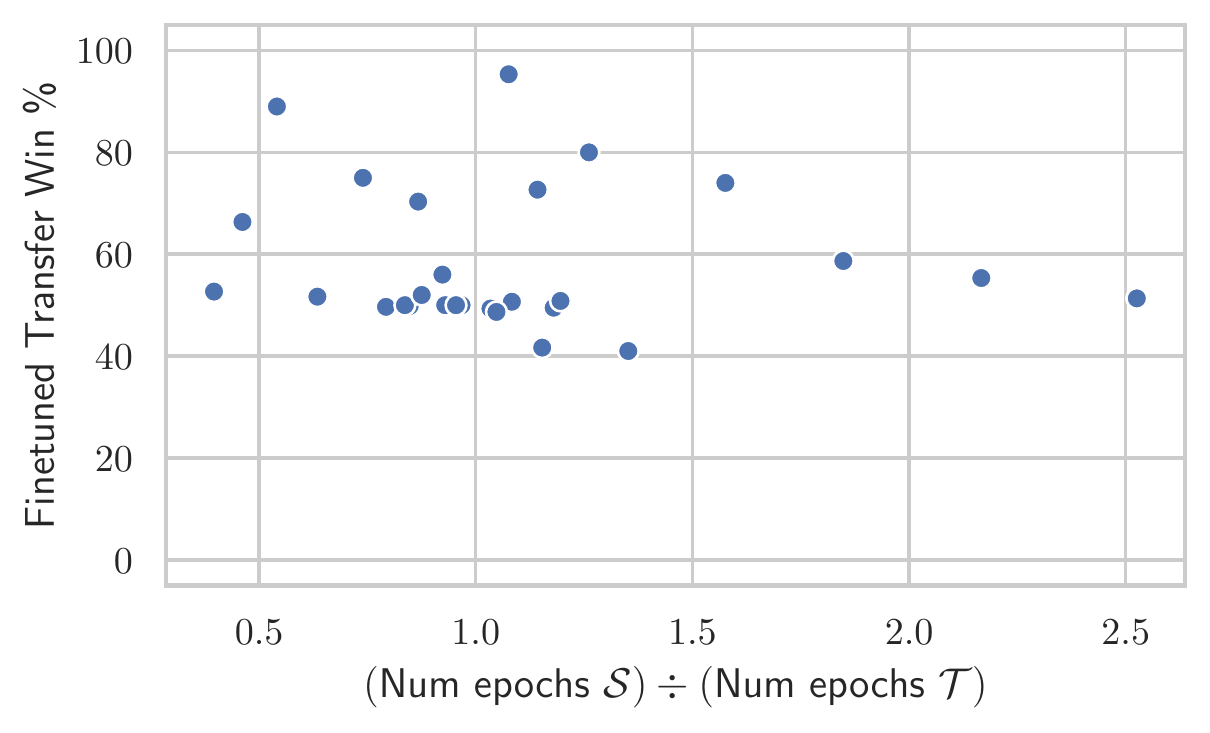}
  \caption{Different board shapes for $\mathcal{S}$ and $\mathcal{T}$.}
  \label{Fig:ResultsFinetuningBoardShapeChanges}
\end{subfigure}
\begin{subfigure}{.34\textwidth}
  \centering
  \includegraphics[width=\linewidth]{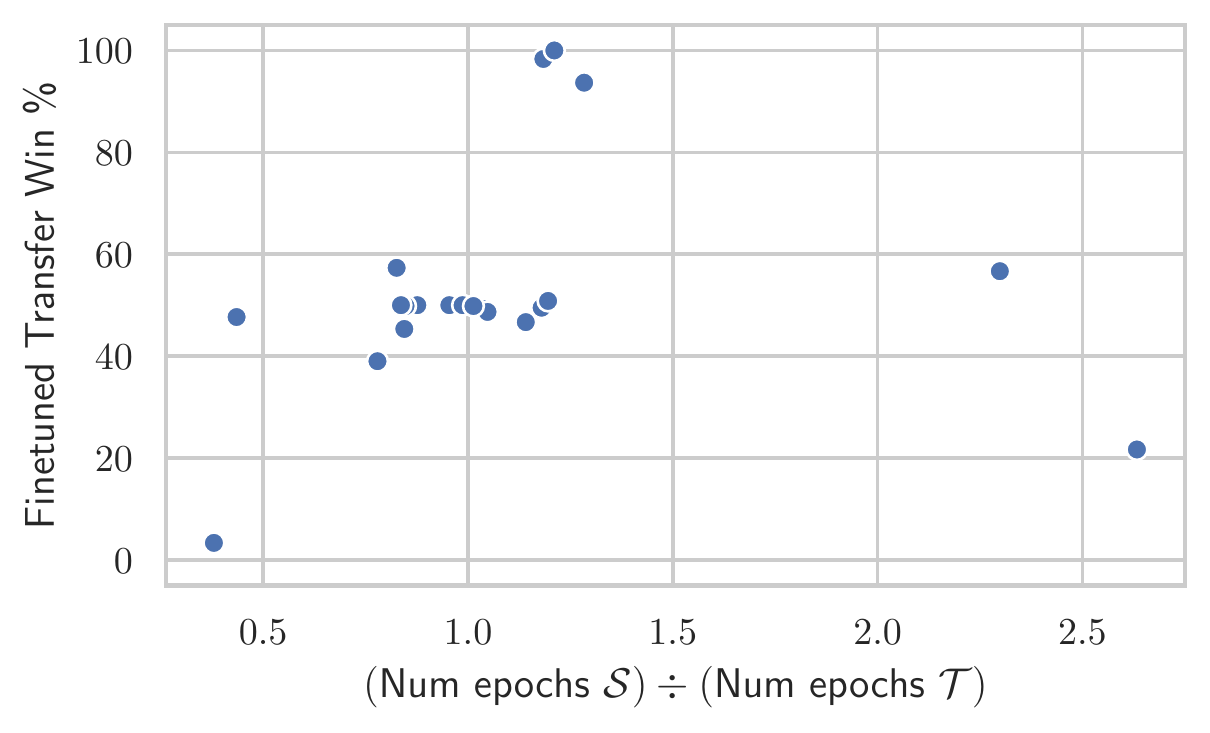}
  \caption{Different win conditions for $\mathcal{S}$ and $\mathcal{T}$.}
  \label{Fig:ResultsFinetuningRuleChanges}
\end{subfigure}
\caption{Win percentages of models trained on $\mathcal{S}$ and subsequently fine-tuned on $\mathcal{T}$, against models trained only on $\mathcal{T}$ -- evaluated on $\mathcal{T}$. 
}
\label{Fig:ResultsFinetuning}
\end{figure*}

For every model transferred from a source domain $\mathcal{S}$ to a target domain $\mathcal{T}$ as described above, we train it under identical conditions as the initial training runs -- for an additional 20 hours -- to evaluate the effect of using transfer for initialisation of the network. \reffigure{Fig:ResultsFinetuning} depicts scatterplots of win percentages for four distinct cases; transfer to variants with smaller board sizes, with larger board sizes, with different board shapes, and with different win conditions. There are many cases of win percentages close to $50\%$, which can be interpreted as cases where transfer neither helped nor hurt final performance, and many cases with higher win percentages -- which can be interpreted as cases where transfer increased final performance in comparison to training from scratch. We observe a small number of cases, especially when $\mathcal{S}$ has a larger board than $\mathcal{T}$, or has different win conditions, in which there is clear negative transfer \cite{Zhang2020NegativeTransfer} and the final performance is still closer to $0\%$ even after fine-tuning on $\mathcal{T}$. More detailed results are provided in Appendix E of the supplementary material.

\subsubsection{Transfer Between Different Games}

For our final set of experiments, we collected four sets of games, and within each set carried out similar experiments as described above -- this time transferring models between distinct games, rather than game variants. The first set consists of six different \textbf{Line Completion Games}; in each of these games the win condition is to create a line of $n$ pieces, but the games differ in aspects such as the value of $n$, board sizes and shapes, move rules, loss conditions, etc. We evaluate transfer from each of those games, to each of these games. The second set consists of four \textbf{Shogi Variants}: we include \textit{Hasami Shogi}, \textit{Kyoto Shogi}, \textit{Minishogi}, and \textit{Shogi}, and evaluate transfer from and to each of them. In the third set we evaluate transfer from each of four variants of \textit{Broken Line}, to each of the six line completion games. \textit{Broken Line} is a custom-made line completion game where only diagonal lines count towards the win condition, whereas the standard line completion games allow for orthogonal lines. In the fourth set, we evaluate transfer from each of five variants of \textit{Diagonal Hex}, to each of six variants of \textit{Hex}. \textit{Diagonal Hex} only considers diagonal connections for the win condition of \textit{Hex}, whereas the \textit{Hex} variants only consider orthogonal connections. Appendix C of the supplementary material provides more details on all the games and variants.

In most cases, zero-shot win percentages for models transferred between distinct games are close to $0\%$. We observe some success with zero-shot win percentages greater than $30\%$ for transfer from several different line completion games to \textit{Connect 6}, zero-shot win percentages between $20\%$ and $50\%$ for transfer from three different Shogi variants to \textit{Hasami Shogi}, as well as a win percentage of $97\%$ for zero-shot transfer from \textit{Minishogi} to \textit{Shogi}. Appendix F of the supplementary material contains more detailed results.

\begin{figure}
    \centering
    \includegraphics[width=.9\linewidth]{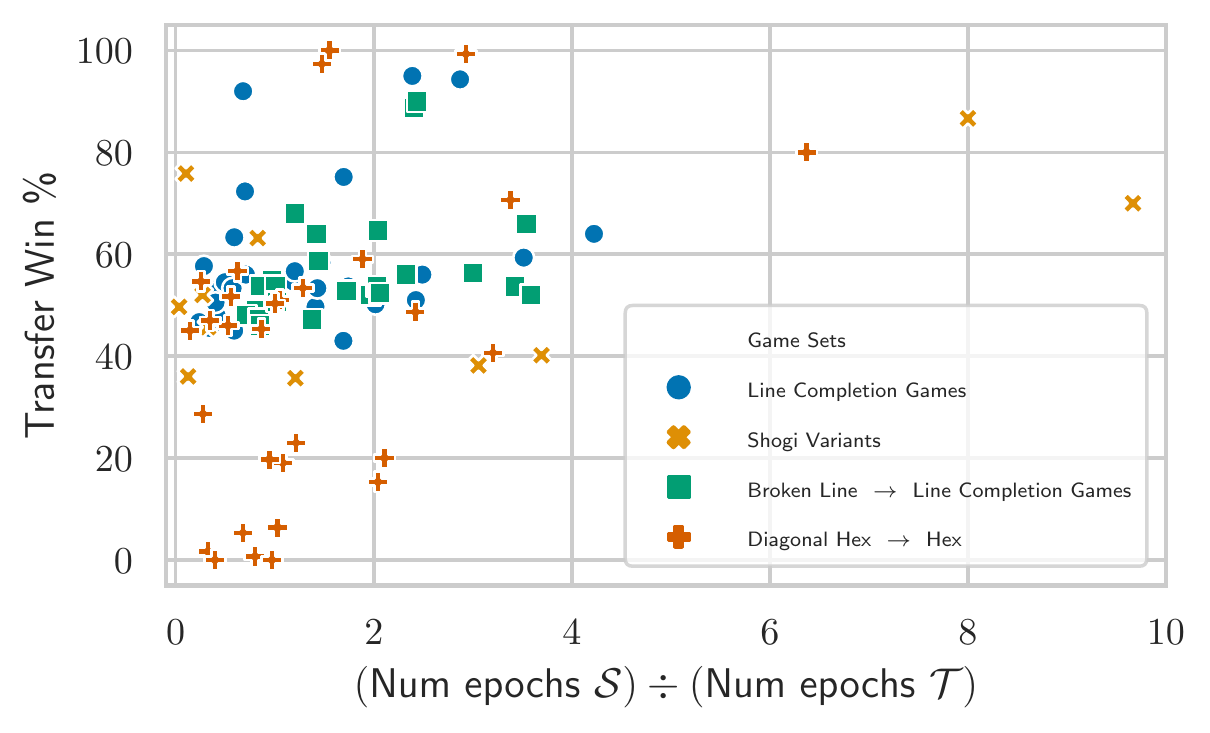}
    \caption{Win percentages of models that were trained in $\mathcal{S}$, transferred to $\mathcal{T}$, and fine-tuned in $\mathcal{T}$, evaluated in $\mathcal{T}$ against models trained directly in $\mathcal{T}$. $\mathcal{S}$ and $\mathcal{T}$ are different games.}
    \label{Fig:ResultsCrossGame}
\end{figure}

\reffigure{Fig:ResultsCrossGame} depicts win percentages for transferred models after they received an additional 20 hours of fine-tuning time on $\mathcal{T}$. Most notably for various cases of transfer from \textit{Diagonal Hex} to \textit{Hex}, there is a high degree of negative transfer, with many win percentages far below $50\%$ even after fine-tuning. It may be that the differences in connectivity rules are too big to allow for consistently successful transfer. The difficulties in transfer may also be due to large differences in the distributions of outcomes, resulting in large mismatches for the value head; ties are a common result in some variants of \textit{Diagonal Hex}, but impossible in \textit{Hex}. In particular for line completion games, transfer appears to be generally successful; there are no severe cases of negative transfer, and a significant amount with strong positive transfer.

\section{Conclusions}

In this paper, we explored the ability to transfer fully convolutional networks with global pooling, trained using an AlphaZero-like approach, between variants of board games, and distinct board games. Firstly, we compared the performance of such architectures to various others outside of a transfer learning setting, and demonstrated them to be among the top-performing architectures in a variety of board games: fully convolutional nets bring a strong improvement in particular for large board games, and global pooling and U-nets provide slight improvements (Fig. \ref{Fig:FullyConvArch}). 
Secondly, we explored how to transfer parameters of such networks between different games with different state and action representations in the Ludii general game system. We evaluated zero-shot transfer performance, as well as the performance of transferred models after additional fine-tuning in the target domain, for a wide variety of source and target games and game variants in Ludii. We find several cases where even zero-shot transfer is highly successful -- especially when transferring from smaller games to larger ones. We also observe a significant number of cases of beneficial transfer after fine-tuning, even when the source and target domains have more significant differences than just changes in board size or shape (see \reffigure{Fig:ResultsCrossGame}).
Finally, we find some cases with clear negative transfer, even after fine-tuning, which point to avenues for future research.

\section*{Acknowledgements}
The authors would like to thank Nicolas Usunier for comments on an earlier version of this work. This work was partially supported by the European Research Council as part of the Digital Ludeme Project (ERC Consolidator Grant \#771292), led by Cameron Browne at Maastricht University’s Department of Data Science and Knowledge Engineering.

\bibliography{bibliobrain,rl,archi}
\bibliographystyle{icml2021}

\clearpage
\onecolumn
\appendix

\section{Directly Transferable Ludii State Channels}

Most state channels in Ludii's tensor representations \cite{Soemers2021DeepLearning} are directly transferable between games, in the sense that they encode semantically similar data for any game in which they are present, and can be modelled as channels that always have only values of $0$ in any game that they are not present in. These channels are briefly described here:
\begin{itemize}
    \item A channel encoding the height of a stack of pieces for every position (only present in games that allow for pieces of more than a single piece type to stack on the same position).
    \item A channel encoding the number of pieces per position (only present in games that allow multiple pieces of the same type to form a pile on the same position).
    \item Channels encoding the (typically monetary) ``amount'' value per player.
    \item Binary channels encoding whether a given player is the current player to move.
    \item Channels encoding ``local state'' values per position (for instance used to memorise whether pieces moved to determine the legality of castling in \textit{Chess}).
    \item A channel encoding whether or not players have swapped roles.
    \item Channels encoding ``from'' and ``to'' positions of the last and second-to-last moves.
\end{itemize}

\section{Details on Experimental Setup} \label{Appendix:Hyperparams}

For all training runs for transfer learning experiments, the following command-line arguments were supplied to the \texttt{train} command of Polygames \cite{Cazenave2020Polygames}:

\begin{itemize}
    \item \verb|--num_game 2|: Affects the number of threads used to run games per self-play client process. 
    \item \verb|--epoch_len 256|: Number of training batches per epoch.
    \item \verb|--batchsize 128|: Batch size for model training.
    \item \verb|--sync_period 32|: Affects how often models are synced.
    \item \verb|--num_rollouts 400|: Number of MCTS iterations per move during self-play training.
    \item \verb|--replay_capacity 100000|: Capacity of replay buffer.
    \item \verb|--replay_warmup 9000|: Minimum size of replay buffer before training starts.
    \item \verb|--model_name "ResConvConvLogitPoolModelV2"|: Type of architecture to use (a fully convolutional architecture with global pooling).
    \item \verb|--bn|: Use of batch normalization \cite{Ioffe2015BatchNormalization}.
    \item \verb|--nnsize 2|: A value of $2$ means that hidden convolutional layers each have twice as many channels as the number of channels for the state input tensors.
    \item \verb|--nb_layers_per_net 6|: Number of convolutional layers per residual block.
    \item \verb|--nb_nets 10|: Number of residual blocks.
    \item \verb|--tournament_mode=true|: Use the tournament mode of Polygames to select checkpoints to play against in self-play.
    \item \verb|--bsfinder_max_bs=800|: Upper bound on number of neural network queries batched together during inference (we used a lower value of $400$ to reduce memory usage in \textit{Breakthrough}, \textit{Hasami Shogi}, \textit{Kyoto Shogi}, \textit{Minishogi}, \textit{Shogi}, and \textit{Tobi Shogi}).
\end{itemize}

All evaluation games in transfer learning experiments were run using the following command-line arguments for the \texttt{eval} command of Polygames:

\begin{itemize}
    \item \verb|--num_actor_eval=1|: Number of threads running simultaneously for a single MCTS search for the agent being evaluated.
    \item \verb|--num_rollouts_eval=800|: Number of MCTS iterations per move for the agent being evaluated.
    \item \verb|--num_actor_opponent=1|: Number of threads running simultaneously for a single MCTS search for the baseline agent.
    \item \verb|--num_rollouts_opponent=800|: Number of MCTS iterations per move for the baseline agent.
\end{itemize}

Any parameters not listed were left at their defaults in the Polygames implementation.

\section{Details on Games and Game Variants}

This section provides additional details on all the games and variants of games used throughout all the experiments described in the paper. A game with name \verb|GameName| is selected in Polygames by providing \verb|--game_name="GameName"| as command-line argument. For games implemented in Ludii, non-default variants are loaded by providing \verb|--game_options "X" "Y" "Z"| as additional command-line arguments, where \verb|X|, \verb|Y|, and \verb|Z| refer to one or more option strings.

\subsection{Polygames Games}

For the evaluation of fully convolutional architectures, we used games as implemented directly in Polygames. \reftable{Table:PolygamesGames} lists the exact game names used. Note that all versions of \textit{Havannah} and \textit{Hex} included use of the pie rule (or swap rule).

\begin{table}[H]
\caption{Game implementations from Polygames used for evaluation of fully convolutional architectures. The right column shows the names used in command-line arguments.}
\label{Table:PolygamesGames}
\vspace{6pt}
\centering
\begin{tabular}{@{}ll@{}}
\toprule
\textbf{Game} & \textbf{Game Name Argument} \\
\midrule
Connect6 & \verb|Connect6| \\
Havannah ($5$$\times$$5$) & \verb|Havannah5pie| \\
Havannah ($8$$\times$$8$) & \verb|Havannah8pie| \\
Havannah ($10$$\times$$10$) & \verb|Havannah10pie| \\
Hex ($11$$\times$$11$) & \verb|Hex11pie| \\
Hex ($13$$\times$$13$) & \verb|Hex13pie| \\
Hex ($19$$\times$$19$) & \verb|Hex19pie| \\
Minishogi & \verb|Minishogi| \\
\bottomrule
\end{tabular}
\end{table}

\subsection{Ludii Game Variants}

For the transfer learning experiments between variants of games, we used nine games -- each with multiple variants -- as implemented in Ludii: \textit{Breakthrough}, \textit{Broken Line}, \textit{Diagonal Hex}, \textit{Gomoku}, \textit{Hex}, \textit{HeXentafl}, \textit{Konane}, \textit{Pentalath}, and \textit{Yavalath}. For each of these games, Tables \ref{Table:BreakthroughDetails}-\ref{Table:YavalathDetails} provide additional details. In each of these tables, the final column lists the number of trainable parameters in the Deep Neural Network (DNN) that is constructed for each game variant, using hyperparameters as described in \refappendix{Appendix:Hyperparams}.

\begin{table}[H]
\caption{Details on \textit{Breakthrough} variants. This implementation of Breakthrough is loaded in Polygames using ``LudiiBreakthrough.lud'' as game name. By default, Breakthrough is played on an $8$$\times$$8$ square board.}
\label{Table:BreakthroughDetails}
\vspace{6pt}
\centering
\begin{tabular}{@{}lllr@{}}
\toprule
\textbf{Variant} & \textbf{Options} & \textbf{Description} & \textbf{Num. Params DNN} \\
\midrule
Square6 & \verb|"Board Size/6x6" "Board/Square"| & $6$$\times$$6$ square board & 188,296 \\
Square8 & \verb|"Board Size/8x8" "Board/Square"| & $8$$\times$$8$ square board & 188,296 \\
Square10 & \verb|"Board Size/10x10" "Board/Square"| & $10$$\times$$10$ square board & 188,296 \\
Hexagon4 & \verb|"Board Size/4x4" "Board/Hexagon"| & $4$$\times$$4$ hexagonal board & 188,296 \\
Hexagon6 & \verb|"Board Size/6x6" "Board/Hexagon"| & $6$$\times$$6$ hexagonal board & 188,296 \\
Hexagon8 & \verb|"Board Size/8x8" "Board/Hexagon"| & $8$$\times$$8$ hexagonal board & 188,296 \\
\bottomrule
\end{tabular}
\end{table}

\begin{table}[H]
\caption{Details on \textit{Broken Line} variants. This implementation of Broken Line is loaded in Polygames using ``LudiiBroken Line.lud'' as game name. }
\label{Table:BrokenLineDetails}
\vspace{6pt}
\centering
\begin{tabular}{@{}lp{5cm}p{5cm}r@{}}
\toprule
\textbf{Variant} & \textbf{Options} & \textbf{Description} & \textbf{Num. Params DNN} \\
\midrule
LineSize3Hex & \verb|"Line Size/3"| \verb|"Board Size/5x5"| \verb|"Board/hex"| & $5$$\times$$5$ hexagonal board, lines of $3$ win & 222,464 \\
LineSize4Hex & \verb|"Line Size/4"| \verb|"Board Size/5x5"| \verb|"Board/hex"| & $5$$\times$$5$ hexagonal board, lines of $4$ win & 222,464 \\
LineSize5Square & \verb|"Line Size/5"| \verb|"Board Size/9x9"| \verb|"Board/Square"| & $9$$\times$$9$ square board, lines of $5$ win & 222,464 \\
LineSize6Square & \verb|"Line Size/6"| \verb|"Board Size/9x9"| \verb|"Board/Square"| & $9$$\times$$9$ square board, lines of $6$ win & 222,464 \\
\bottomrule
\end{tabular}
\end{table}

\begin{table}[H]
\caption{Details on \textit{Diagonal Hex} variants. This implementation of Diagonal Hex is loaded in Polygames using ``LudiiDiagonal Hex.lud'' as game name. }
\label{Table:DiagonalHexDetails}
\vspace{6pt}
\centering
\begin{tabular}{@{}lllr@{}}
\toprule
\textbf{Variant} & \textbf{Options} & \textbf{Description} & \textbf{Num. Params DNN} \\
\midrule
$7$$\times$$7$ & \verb|"Board Size/7x7"| & $7$$\times$$7$ hexagonal board & 222,464 \\
$9$$\times$$9$ & \verb|"Board Size/9x9"| & $9$$\times$$9$ hexagonal board & 222,464 \\
$11$$\times$$11$ & \verb|"Board Size/11x11"| & $11$$\times$$11$ square board & 222,464 \\
$13$$\times$$13$ & \verb|"Board Size/13x13"| & $13$$\times$$13$ square board & 222,464 \\
$19$$\times$$19$ & \verb|"Board Size/19x19"| & $19$$\times$$19$ square board & 222,464 \\
\bottomrule
\end{tabular}
\end{table}

\begin{table}[H]
\caption{Details on \textit{Gomoku} variants. This implementation of Gomoku is loaded in Polygames using ``LudiiGomoku.lud'' as game name. By default, Gomoku is played on a $15$$\times$$15$ board. }
\label{Table:GomokuDetails}
\vspace{6pt}
\centering
\begin{tabular}{@{}lllr@{}}
\toprule
\textbf{Variant} & \textbf{Options} & \textbf{Description} & \textbf{Num. Params DNN} \\
\midrule
$9$$\times$$9$ & \verb|"Board Size/9x9"| & $9$$\times$$9$ square board & 180,472 \\
$13$$\times$$13$ & \verb|"Board Size/13x13"| & $13$$\times$$13$ square board & 180,472 \\
$15$$\times$$15$ & \verb|"Board Size/15x15"| & $15$$\times$$15$ square board & 180,472 \\
$19$$\times$$19$ & \verb|"Board Size/19x19"| & $19$$\times$$19$ square board & 180,472 \\
\bottomrule
\end{tabular}
\end{table}

\begin{table}[H]
\caption{Details on \textit{Hex} variants. This implementation of Hex is loaded in Polygames using ``LudiiHex.lud'' as game name. By default, Hex is played on an $11$$\times$$11$ board. }
\label{Table:HexDetails}
\vspace{6pt}
\centering
\begin{tabular}{@{}lp{4cm}p{6cm}r@{}}
\toprule
\textbf{Variant} & \textbf{Options} & \textbf{Description} & \textbf{Num. Params DNN} \\
\midrule
$7$$\times$$7$ & \verb|"Board Size/7x7"| & $7$$\times$$7$ board, standard win condition & 222,464 \\
$9$$\times$$9$ & \verb|"Board Size/9x9"| & $9$$\times$$9$ board, standard win condition & 222,464 \\
$11$$\times$$11$ & \verb|"Board Size/11x11"| & $11$$\times$$11$ board, standard win condition & 222,464 \\
$13$$\times$$13$ & \verb|"Board Size/13x13"| & $13$$\times$$13$ board, standard win condition & 222,464 \\
$19$$\times$$19$ & \verb|"Board Size/19x19"| & $19$$\times$$19$ board, standard win condition & 222,464 \\
$11$$\times$$11$ Misere & \verb|"Board Size/11x11"| \verb|"End Rules/Misere"| & $11$$\times$$11$ board, inverted win condition & 222,464 \\
\bottomrule
\end{tabular}
\end{table}

\begin{table}[H]
\caption{Details on \textit{HeXentafl} variants. This implementation of HeXentafl is loaded in Polygames using ``LudiiHeXentafl.lud'' as game name. By default, HeXentafl is played on a $4$$\times$$4$ board. }
\label{Table:HeXentaflDetails}
\vspace{6pt}
\centering
\begin{tabular}{@{}lllr@{}}
\toprule
\textbf{Variant} & \textbf{Options} & \textbf{Description} & \textbf{Num. Params DNN} \\
\midrule
$4$$\times$$4$ & \verb|"Board Size/4x4"| & $4$$\times$$4$ hexagonal board & 231,152 \\
$5$$\times$$5$ & \verb|"Board Size/5x5"| & $5$$\times$$5$ hexagonal board & 231,152 \\
\bottomrule
\end{tabular}
\end{table}

\begin{table}[H]
\caption{Details on \textit{Konane} variants. This implementation of Konane is loaded in Polygames using ``LudiiKonane.lud'' as game name. By default, Konane is played on an $8$$\times$$8$ board. }
\label{Table:KonaneDetails}
\vspace{6pt}
\centering
\begin{tabular}{@{}lllr@{}}
\toprule
\textbf{Variant} & \textbf{Options} & \textbf{Description} & \textbf{Num. Params DNN} \\
\midrule
$6$$\times$$6$ & \verb|"Board Size/6x6"| & $6$$\times$$6$ square board & 188,296 \\
$8$$\times$$8$ & \verb|"Board Size/8x8"| & $8$$\times$$8$ square board & 188,296 \\
$10$$\times$$10$ & \verb|"Board Size/10x10"| & $10$$\times$$10$ square board & 188,296 \\
$12$$\times$$12$ & \verb|"Board Size/12x12"| & $12$$\times$$12$ square board & 188,296 \\
\bottomrule
\end{tabular}
\end{table}

\begin{table}[H]
\caption{Details on \textit{Pentalath} variants. This implementation of Pentalath is loaded in Polygames using ``LudiiPentalath.lud'' as game name. By default, Pentalath is played on half a hexagonal board. }
\label{Table:PentalathDetails}
\vspace{6pt}
\centering
\begin{tabular}{@{}lllr@{}}
\toprule
\textbf{Variant} & \textbf{Options} & \textbf{Description} & \textbf{Num. Params DNN} \\
\midrule
HexHexBoard & \verb|"Board/HexHexBoard"| & A full hexagonal board & 180,472 \\
HalfHexHexBoard & \verb|"Board/HalfHexHexBoard"| & Half a hexagonal board & 180,472 \\
\bottomrule
\end{tabular}
\end{table}

\begin{table}[H]
\caption{Details on \textit{Yavalath} variants. This implementation of Yavalath is loaded in Polygames using ``LudiiYavalath.lud'' as game name. By default, Yavalath is played on a $5$$\times$$5$ board. }
\label{Table:YavalathDetails}
\vspace{6pt}
\centering
\begin{tabular}{@{}lllr@{}}
\toprule
\textbf{Variant} & \textbf{Options} & \textbf{Description} & \textbf{Num. Params DNN} \\
\midrule
$3$$\times$$3$ & \verb|"Board Size/3x3"| & $3$$\times$$3$ hexagonal board & 222,464 \\
$4$$\times$$4$ & \verb|"Board Size/4x4"| & $4$$\times$$4$ hexagonal board & 222,464 \\
$5$$\times$$5$ & \verb|"Board Size/5x5"| & $5$$\times$$5$ hexagonal board & 222,464 \\
$6$$\times$$6$ & \verb|"Board Size/6x6"| & $6$$\times$$6$ hexagonal board & 222,464 \\
$7$$\times$$7$ & \verb|"Board Size/7x7"| & $7$$\times$$7$ hexagonal board & 222,464 \\
$8$$\times$$8$ & \verb|"Board Size/8x8"| & $8$$\times$$8$ hexagonal board & 222,464 \\
\bottomrule
\end{tabular}
\end{table}

\subsection{Ludii Line Completion Games}

For the evaluation of transfer between different line completion games, we used six different line completion games: \textit{Connect6}, \textit{Dai Hasami Shogi}, \textit{Gomoku}, \textit{Pentalath}, \textit{Squava}, and \textit{Yavalath}. Several properties of these games are listed in \reftable{Table:LineCompletionGames}.

\begin{table}[H]
\caption{Details on different line completion games. }
\label{Table:LineCompletionGames}
\vspace{6pt}
\centering
\begin{tabular}{@{}lrrrrrr@{}}
\toprule
& Connect6 & Dai Hasami Shogi & Gomoku & Pentalath & Squava & Yavalath \\
\midrule
Board Shape & Square & Square & Square & Hexagonal & Square & Hexagonal \\
Board Size & $19$$\times$$19$ & $9$$\times$$9$ & $9$$\times$$9$ & $5$$\times$$5$ & $5$$\times$$5$ & $5$$\times$$5$ \\
Win Line Length & $6$ & $5$ & $5$ & $5$ & $4$ & $4$ \\
Loss Line Length & - & - & - & - & $3$ & $3$ \\
Max Win Line Length & - & - & $5$ & - & - & - \\
Can Move Pieces? & $\times$ & \checkmark & $\times$ & $\times$ & $\times$ & $\times$ \\
Can Capture Pieces? & $\times$ & \checkmark & $\times$ & \checkmark & $\times$ & $\times$ \\
Uses Swap Rule? & $\times$ & $\times$ & $\times$ & $\times$ & \checkmark & \checkmark \\
Moves per Turn & $2$* & 1 & 1 & 1 & 1 & 1 \\
State Tensor Shape & $(9, 19, 19)$ & $(9, 9, 9)$ & $(9, 9, 9)$ & $(9, 9, 17)$ & $(10, 5, 5)$ & $(10, 9, 17)$ \\
Policy Tensor Shape & $(3, 19, 19)$ & $(51, 9, 9)$ & $(3, 9, 9)$ & $(3, 9, 17)$ & $(3, 5, 5)$ & $(3, 9, 17)$ \\
Num. Params DNN & 180,472 & 188,296 & 180,472 & 180,472 & 222,464 & 222,464 \\
\midrule
\multicolumn{7}{l}{*The first turn in Connect6 consists of only $1$ move.} \\
\bottomrule
\end{tabular}
\end{table}

\subsection{Ludii Shogi Games}

For the evaluation of transfer between different variants of Shogi, we used four games: \textit{Hasami Shogi}, \textit{Kyoto Shogi}, \textit{Minishogi}, and \textit{Shogi}. Several properties of these games are listed in \reftable{Table:ShogiGames}.

\begin{table}[H]
\caption{Details on variants of Shogi. }
\label{Table:ShogiGames}
\vspace{6pt}
\centering
\begin{tabular}{@{}lrrrr@{}}
\toprule
& Hasami Shogi & Kyoto Shogi & Minishogi & Shogi \\
\midrule
Board Size & $9$$\times$$9$ & $5$$\times$$5$ & $5$$\times$$5$ & $9$$\times$$9$ \\
Num. Piece Types per Player & $1$ & $9$ & $10$ & $14$ \\
Can Drop Captured Pieces? & $\times$ & \checkmark & \checkmark & \checkmark \\
State Tensor Shape & $(9, 9, 9)$ & $(28, 8, 5)$ & $(30, 8, 5)$ & $(38, 12, 9)$ \\
Policy Tensor Shape & $(51, 9, 9)$ & $(51, 8, 5)$ & $(51, 8, 5)$ & $(51, 12, 9)$ \\
Num. Params DNN & 188,296 & 1,752,908 & 2,009,752 & 3,212,648 \\
\bottomrule
\end{tabular}
\end{table}

\subsection{Broken Line and Diagonal Hex}

\begin{figure}[H]
\centering
\hfill
\begin{subfigure}{.3\textwidth}
  \centering
  \includegraphics[width=\linewidth]{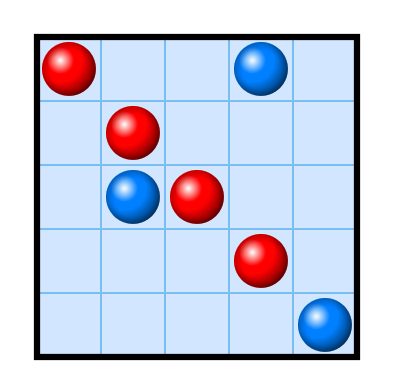}
  \caption{A diagonal line of $4$ on the square board is a win for the red player in \textit{Broken Line}.}
  \label{Fig:BrokenLineSquare}
\end{subfigure}
\hfill
\begin{subfigure}{.3\textwidth}
  \centering
  \includegraphics[width=\linewidth]{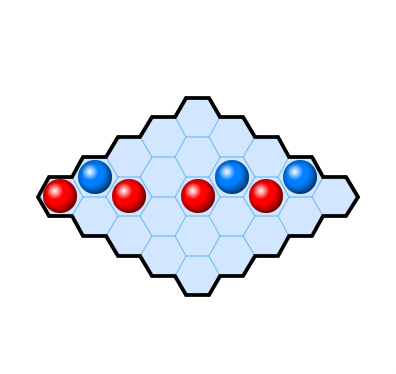}
  \caption{A ``diagonal'' line of $4$ on the hexagonal board is a win for the red player in \textit{Broken Line}.}
  \label{Fig:BrokenLineHexagonal}
\end{subfigure}
\hfill
\begin{subfigure}{.3\textwidth}
  \centering
  \includegraphics[width=\linewidth]{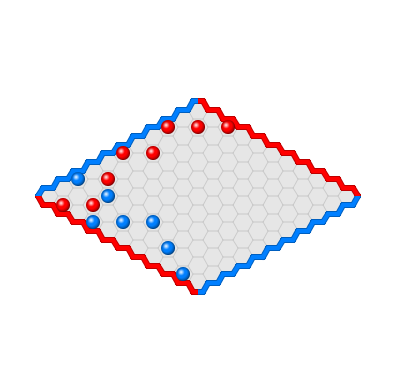}
  \caption{A chain of ``diagonally'' connected pieces on the hexagonal board is a win for the red player in \textit{Diagonal Hex}.}
  \label{Fig:DiagonalHex}
\end{subfigure}
\caption{Examples of winning game states for the red player in \textit{Broken Line} (on a square and hexagonal board), and \textit{Diagonal Hex}. In both examples for \textit{Broken Line}, the target line length was set to $4$.}
\label{Fig:BrokenLineDiagonalHex}
\hfill
\end{figure}

\textit{Broken Line} and \textit{Diagonal Hex} are variations on line completion games, and \textit{Hex}, respectively, which only take into consideration diagonal connections for the line completion and connection win conditions. On hexagonal grids, two cells are considered to be ``diagonally connected'' if there exists an edge that connects exactly one vertex of each of the cells. \reffigure{Fig:BrokenLineDiagonalHex} depicts examples of winning game states for the red player in Broken Line on a square board, Broken Line on a hexagonal board, and Diagonal Hex.

\section{Detailed Results -- Zero-shot Transfer Between Game Variants}

Tables \ref{Table:BreakthroughBoardsZeroShot}-\ref{Table:YavalathBoardsZeroShot} provide detailed results for all evaluations of  zero-shot transfer between variants within each out of nine different games.

\begin{table}[H]
\caption{Win percentage of MCTS with final checkpoint from source domain against MCTS with final checkpoint trained in target domain, evaluated in target domain (zero-shot transfer). Source and target domains are different boards in Breakthrough.}
\label{Table:BreakthroughBoardsZeroShot}
\vspace{6pt}
\centering
\begin{tabular}{@{}lrrrrrr@{}}
\toprule
\textbf{Game: Breakthrough} & \multicolumn{6}{c}{Target Domain} \\
\cmidrule(lr){2-7}
Source Domain & Square6 & Square8 & Square10 & Hexagon4 & Hexagon6 & Hexagon8 \\
\midrule
Square6 & - & 0.00\% & 7.33\% & 0.00\% & 0.00\% & 0.67\% \\
Square8 & 10.00\% & - & 77.00\% & 2.67\% & 0.00\% & 1.00\% \\
Square10 & 1.33\% & 0.33\% & - & 0.67\% & 0.00\% & 0.33\% \\
Hexagon4 & 0.00\% & 0.00\% & 0.00\% & - & 0.33\% & 1.33\% \\
Hexagon6 & 0.67\% & 0.00\% & 0.00\% & 12.67\% & - & 39.67\% \\
Hexagon8 & 0.00\% & 0.00\% & 0.00\% & 4.00\% & 5.00\% & - \\
\bottomrule
\end{tabular}
\end{table}

\begin{table}[H]
\caption{Win percentage of MCTS with final checkpoint from source domain against MCTS with final checkpoint trained in target domain, evaluated in target domain (zero-shot transfer). Source and target domains are different boards in Broken Line.}
\label{Table:BrokenLineBoardsZeroShot}
\vspace{6pt}
\centering
\begin{tabular}{@{}lrrrr@{}}
\toprule
\textbf{Game: Broken Line} & \multicolumn{4}{c}{Target Domain} \\
\cmidrule(lr){2-5}
Source Domain & LineSize3Hex & LineSize4Hex & LineSize5Square & LineSize6Square \\
\midrule
LineSize3Hex & - & 5.67\% & 0.00\% & 0.00\% \\
LineSize4Hex & 19.33\% & - & 0.00\% & 0.17\% \\
LineSize5Square & 7.00\% & 0.00\% & - & 49.67\% \\
LineSize6Square & 3.67\% & 0.00\% & 47.17\% & - \\
\bottomrule
\end{tabular}
\end{table}

\begin{table}[H]
\caption{Win percentage of MCTS with final checkpoint from source domain against MCTS with final checkpoint trained in target domain, evaluated in target domain (zero-shot transfer). Source and target domains are different boards in Diagonal Hex.}
\label{Table:DiagonalHexBoardsZeroShot}
\vspace{6pt}
\centering
\begin{tabular}{@{}lrrrrr@{}}
\toprule
\textbf{Game: Diagonal Hex} & \multicolumn{5}{c}{Target Domain} \\
\cmidrule(lr){2-6}
Source Domain & $7$$\times$$7$ & $9$$\times$$9$ & $11$$\times$$11$ & $13$$\times$$13$ & $19$$\times$$19$ \\
\midrule
$7$$\times$$7$ & - & 38.00\% & 22.50\% & 100.00\% & 99.67\% \\
$9$$\times$$9$ & 45.17\% & - & 83.00\% & 100.00\% & 100.00\% \\
$11$$\times$$11$ & 13.00\% & 18.00\% & - & 100.00\% & 100.00\% \\
$13$$\times$$13$ & 0.00\% & 0.00\% & 0.00\% & - & 44.83\% \\
$19$$\times$$19$ & 0.00\% & 0.00\% & 0.00\% & 33.83\% & - \\
\bottomrule
\end{tabular}
\end{table}

\begin{table}[H]
\caption{Win percentage of MCTS with final checkpoint from source domain against MCTS with final checkpoint trained in target domain, evaluated in target domain (zero-shot transfer). Source and target domains are different boards in Gomoku.}
\label{Table:GomokuBoardsZeroShot}
\vspace{6pt}
\centering
\begin{tabular}{@{}lrrrr@{}}
\toprule
\textbf{Game: Gomoku} & \multicolumn{4}{c}{Target Domain} \\
\cmidrule(lr){2-5}
Source Domain & $9$$\times$$9$ & $13$$\times$$13$ & $15$$\times$$15$ & $19$$\times$$19$ \\
\midrule
$9$$\times$$9$ & - & 44.00\% & 31.67\% & 18.67\% \\
$13$$\times$$13$ & 28.17\% & - & 51.33\% & 62.00\% \\
$15$$\times$$15$ & 25.50\% & 40.50\% & - & 66.00\% \\
$19$$\times$$19$ & 19.83\% & 32.67\% & 35.67\% & - \\
\bottomrule
\end{tabular}
\end{table}

\begin{table}[H]
\caption{Win percentage of MCTS with final checkpoint from source domain against MCTS with final checkpoint trained in target domain, evaluated in target domain (zero-shot transfer). Source and target domains are different variants of Hex.}
\label{Table:HexBoardsZeroShot}
\vspace{6pt}
\centering
\begin{tabular}{@{}lrrrrrr@{}}
\toprule
\textbf{Game: Hex} & \multicolumn{6}{c}{Target Domain} \\
\cmidrule(lr){2-7}
Source Domain & $7$$\times$$7$ & $9$$\times$$9$ & $11$$\times$$11$ & $13$$\times$$13$ & $19$$\times$$19$ & $11$$\times$$11$ Misere \\
\midrule
$7$$\times$$7$ & - & 38.33\% & 14.67\% & 76.67\% & 91.67\% & 0.00\% \\
$9$$\times$$9$ & 21.67\% & - & 56.33\% & 100.00\% & 100.00\% & 0.00\% \\
$11$$\times$$11$ & 20.33\% & 30.33\% & - & 100.00\% & 100.00\% & 0.00\% \\
$13$$\times$$13$ & 4.67\% & 0.67\% & 0.00\% & - & 100.00\% & 0.00\% \\
$19$$\times$$19$ & 0.00\% & 0.00\% & 0.00\% & 0.00\% & - & 0.00\% \\
$11$$\times$$11$ Misere & 0.00\% & 0.00\% & 0.00\% & 0.00\% & 0.00\% & - \\
\bottomrule
\end{tabular}
\end{table}

\begin{table}[H]
\caption{Win percentage of MCTS with final checkpoint from source domain against MCTS with final checkpoint trained in target domain, evaluated in target domain (zero-shot transfer). Source and target domains are different boards in HeXentafl.}
\label{Table:HeXentaflBoardsZeroShot}
\vspace{6pt}
\centering
\begin{tabular}{@{}lrr@{}}
\toprule
\textbf{Game: HeXentafl} & \multicolumn{2}{c}{Target Domain} \\
\cmidrule(lr){2-3}
Source Domain & $4$$\times$$4$ & $5$$\times$$5$ \\
\midrule
$4$$\times$$4$ & - & 15.50\% \\
$5$$\times$$5$ & 9.67\% & - \\
\bottomrule
\end{tabular}
\end{table}

\begin{table}[H]
\caption{Win percentage of MCTS with final checkpoint from source domain against MCTS with final checkpoint trained in target domain, evaluated in target domain (zero-shot transfer). Source and target domains are different boards in Konane.}
\label{Table:KonaneBoardsZeroShot}
\vspace{6pt}
\centering
\begin{tabular}{@{}lrrrr@{}}
\toprule
\textbf{Game: Konane} & \multicolumn{4}{c}{Target Domain} \\
\cmidrule(lr){2-5}
Source Domain & $6$$\times$$6$ & $8$$\times$$8$ & $10$$\times$$10$ & $12$$\times$$12$ \\
\midrule
$6$$\times$$6$ & - & 3.00\% & 14.67\% & 63.33\% \\
$8$$\times$$8$ & 31.00\% & - & 94.00\% & 100.00\% \\
$10$$\times$$10$ & 12.00\% & 3.33\% & - & 99.67\% \\
$12$$\times$$12$ & 8.00\% & 0.00\% & 2.00\% & - \\
\bottomrule
\end{tabular}
\end{table}

\begin{table}[H]
\caption{Win percentage of MCTS with final checkpoint from source domain against MCTS with final checkpoint trained in target domain, evaluated in target domain (zero-shot transfer). Source and target domains are different boards in Pentalath.}
\label{Table:PentalathBoardsZeroShot}
\vspace{6pt}
\centering
\begin{tabular}{@{}lrr@{}}
\toprule
\textbf{Game: Pentalath} & \multicolumn{2}{c}{Target Domain} \\
\cmidrule(lr){2-3}
Source Domain & HexHexBoard & HalfHexHexBoard \\
\midrule
HexHexBoard & - & 26.67\% \\
HalfHexHexBoard & 18.00\% & - \\
\bottomrule
\end{tabular}
\end{table}

\begin{table}[H]
\caption{Win percentage of MCTS with final checkpoint from source domain against MCTS with final checkpoint trained in target domain, evaluated in target domain (zero-shot transfer). Source and target domains are different boards in Yavalath.}
\label{Table:YavalathBoardsZeroShot}
\vspace{6pt}
\centering
\begin{tabular}{@{}lrrrrrr@{}}
\toprule
\textbf{Game: Yavalath} & \multicolumn{6}{c}{Target Domain} \\
\cmidrule(lr){2-7}
Source Domain & $3$$\times$$3$ & $4$$\times$$4$ & $5$$\times$$5$ & $6$$\times$$6$ & $7$$\times$$7$ & $8$$\times$$8$ \\
\midrule
$3$$\times$$3$ & - & 10.83\% & 4.33\% & 1.67\% & 0.67\% & 0.33\% \\
$4$$\times$$4$ & 29.83\% & - & 29.83\% & 15.33\% & 7.67\% & 4.33\% \\
$5$$\times$$5$ & 10.33\% & 12.17\% & - & 30.33\% & 34.00\% & 25.00\% \\
$6$$\times$$6$ & 8.17\% & 20.17\% & 41.17\% & - & 45.33\% & 58.00\% \\
$7$$\times$$7$ & 8.50\% & 21.00\% & 33.00\% & 38.00\% & - & 53.33\% \\
$8$$\times$$8$ & 7.67\% & 13.00\% & 31.00\% & 29.83\% & 47.67\% & - \\
\bottomrule
\end{tabular}
\end{table}

\section{Detailed Results -- Transfer Between Game Variants With Fine-tuning}

Tables \ref{Table:BreakthroughBoardsFinetuned}-\ref{Table:YavalathBoardsFinetuned} provide detailed results for all evaluations of transfer performance after fine-tuning, for transfer between variants within each out of nine different games. Models are trained for 20 hours on the source domain, followed by 20 hours on the target domain, and evaluated against models trained for 20 hours only on the target domain. Tables \ref{Table:BreakthroughBoardsFinetunedReinit}-\ref{Table:YavalathBoardsFinetunedReinit} provide additional results for a similar evaluation where we reinitialised all the parameters of the final convolutional layers before policy and value heads prior to fine-tuning. The basic idea behind this experiment was that it would lead to a more random, less biased policy generating experience from self-play at the start of a fine-tuning process, and hence may improve fine-tuning transfer in cases where full transfer produces a poor initial policy. Overall we did not observe many major changes in transfer performance.

\begin{table}[H]
\caption{Win percentage of MCTS with final checkpoint from source domain against MCTS with final checkpoint trained in target domain, evaluated in target domain after fine-tuning. Source and target domains are different boards in Breakthrough.}
\label{Table:BreakthroughBoardsFinetuned}
\vspace{6pt}
\centering
\begin{tabular}{@{}lrrrrrr@{}}
\toprule
\textbf{Game: Breakthrough} & \multicolumn{6}{c}{Target Domain} \\
\cmidrule(lr){2-7}
Source Domain & Square6 & Square8 & Square10 & Hexagon4 & Hexagon6 & Hexagon8 \\
\midrule
Square6 & - & 87.33\% & 99.00\% & 50.67\% & 74.00\% & 51.33\% \\
Square8 & 50.33\% & - & 92.67\% & 50.00\% & 41.00\% & 55.33\% \\
Square10 & 52.00\% & 64.33\% & - & 49.67\% & 41.67\% & 58.67\% \\
Hexagon4 & 56.00\% & 95.33\% & 80.00\% & - & 74.67\% & 56.67\% \\
Hexagon6 & 51.67\% & 75.00\% & 70.33\% & 50.00\% & - & 74.33\% \\
Hexagon8 & 52.67\% & 66.33\% & 89.00\% & 49.00\% & 74.67\% & - \\
\bottomrule
\end{tabular}
\end{table}

\begin{table}[H]
\caption{Win percentage of MCTS with final checkpoint from source domain against MCTS with final checkpoint trained in target domain, evaluated in target domain after fine-tuning. Source and target domains are different boards in Broken Line.}
\label{Table:BrokenLineBoardsFinetuned}
\vspace{6pt}
\centering
\begin{tabular}{@{}lrrrr@{}}
\toprule
\textbf{Game: Broken Line} & \multicolumn{4}{c}{Target Domain} \\
\cmidrule(lr){2-5}
Source Domain & LineSize3Hex & LineSize4Hex & LineSize5Square & LineSize6Square \\
\midrule
LineSize3Hex & - & 46.67\% & 50.00\% & 50.00\% \\
LineSize4Hex & 50.00\% & - & 49.83\% & 50.00\% \\
LineSize5Square & 49.33\% & 49.50\% & - & 50.00\% \\
LineSize6Square & 48.67\% & 50.83\% & 49.83\% & - \\
\bottomrule
\end{tabular}
\end{table}

\begin{table}[H]
\caption{Win percentage of MCTS with final checkpoint from source domain against MCTS with final checkpoint trained in target domain, evaluated in target domain after fine-tuning. Source and target domains are different boards in Diagonal Hex.}
\label{Table:DiagonalHexBoardsFinetuned}
\vspace{6pt}
\centering
\begin{tabular}{@{}lrrrrr@{}}
\toprule
\textbf{Game: Diagonal Hex} & \multicolumn{5}{c}{Target Domain} \\
\cmidrule(lr){2-6}
Source Domain & $7$$\times$$7$ & $9$$\times$$9$ & $11$$\times$$11$ & $13$$\times$$13$ & $19$$\times$$19$ \\
\midrule
$7$$\times$$7$ & - & 48.83\% & 84.50\% & 100.00\% & 100.00\% \\
$9$$\times$$9$ & 54.33\% & - & 86.00\% & 100.00\% & 100.00\% \\
$11$$\times$$11$ & 54.00\% & 45.50\% & - & 100.00\% & 100.00\% \\
$13$$\times$$13$ & 55.00\% & 49.67\% & 12.83\% & - & 41.00\% \\
$19$$\times$$19$ & 54.17\% & 48.50\% & 0.00\% & 42.50\% & - \\
\bottomrule
\end{tabular}
\end{table}

\begin{table}[H]
\caption{Win percentage of MCTS with final checkpoint from source domain against MCTS with final checkpoint trained in target domain, evaluated in target domain after fine-tuning. Source and target domains are different boards in Gomoku.}
\label{Table:GomokuBoardsFinetuned}
\vspace{6pt}
\centering
\begin{tabular}{@{}lrrrr@{}}
\toprule
\textbf{Game: Gomoku} & \multicolumn{4}{c}{Target Domain} \\
\cmidrule(lr){2-5}
Source Domain & $9$$\times$$9$ & $13$$\times$$13$ & $15$$\times$$15$ & $19$$\times$$19$ \\
\midrule
$9$$\times$$9$ & - & 69.50\% & 70.33\% & 78.00\% \\
$13$$\times$$13$ & 54.33\% & - & 64.67\% & 72.67\% \\
$15$$\times$$15$ & 62.17\% & 50.50\% & - & 67.33\% \\
$19$$\times$$19$ & 55.50\% & 66.67\% & 64.50\% & - \\
\bottomrule
\end{tabular}
\end{table}

\begin{table}[H]
\caption{Win percentage of MCTS with final checkpoint from source domain against MCTS with final checkpoint trained in target domain, evaluated in target domain after fine-tuning. Source and target domains are different variants of Hex.}
\label{Table:HexBoardsFinetuned}
\vspace{6pt}
\centering
\begin{tabular}{@{}lrrrrrr@{}}
\toprule
\textbf{Game: Hex} & \multicolumn{6}{c}{Target Domain} \\
\cmidrule(lr){2-7}
Source Domain & $7$$\times$$7$ & $9$$\times$$9$ & $11$$\times$$11$ & $13$$\times$$13$ & $19$$\times$$19$ & $11$$\times$$11$ Misere \\
\midrule
$7$$\times$$7$ & - & 69.33\% & 81.33\% & 100.00\% & 100.00\% & 56.67\% \\
$9$$\times$$9$ & 47.67\% & - & 73.00\% & 100.00\% & 100.00\% & 93.67\% \\
$11$$\times$$11$ & 48.33\% & 71.00\% & - & 100.00\% & 100.00\% & 98.33\% \\
$13$$\times$$13$ & 47.67\% & 27.67\% & 40.00\% & - & 100.00\% & 57.33\% \\
$19$$\times$$19$ & 50.00\% & 48.33\% & 44.67\% & 82.33\% & - & 3.33\% \\
$11$$\times$$11$ Misere & 47.67\% & 39.00\% & 45.33\% & 100.00\% & 21.67\% & - \\
\bottomrule
\end{tabular}
\end{table}

\begin{table}[H]
\caption{Win percentage of MCTS with final checkpoint from source domain against MCTS with final checkpoint trained in target domain, evaluated in target domain after fine-tuning. Source and target domains are different boards in HeXentafl.}
\label{Table:HeXentaflBoardsFinetuned}
\vspace{6pt}
\centering
\begin{tabular}{@{}lrr@{}}
\toprule
\textbf{Game: HeXentafl} & \multicolumn{2}{c}{Target Domain} \\
\cmidrule(lr){2-3}
Source Domain & $4$$\times$$4$ & $5$$\times$$5$ \\
\midrule
$4$$\times$$4$ & - & 50.50\% \\
$5$$\times$$5$ & 55.83\% & - \\
\bottomrule
\end{tabular}
\end{table}

\begin{table}[H]
\caption{Win percentage of MCTS with final checkpoint from source domain against MCTS with final checkpoint trained in target domain, evaluated in target domain after fine-tuning. Source and target domains are different boards in Konane.}
\label{Table:KonaneBoardsFinetuned}
\vspace{6pt}
\centering
\begin{tabular}{@{}lrrrr@{}}
\toprule
\textbf{Game: Konane} & \multicolumn{4}{c}{Target Domain} \\
\cmidrule(lr){2-5}
Source Domain & $6$$\times$$6$ & $8$$\times$$8$ & $10$$\times$$10$ & $12$$\times$$12$ \\
\midrule
$6$$\times$$6$ & - & 52.33\% & 65.00\% & 99.67\% \\
$8$$\times$$8$ & 51.00\% & - & 94.67\% & 98.33\% \\
$10$$\times$$10$ & 49.33\% & 62.00\% & - & 100.00\% \\
$12$$\times$$12$ & 52.00\% & 20.67\% & 56.67\% & - \\
\bottomrule
\end{tabular}
\end{table}

\begin{table}[H]
\caption{Win percentage of MCTS with final checkpoint from source domain against MCTS with final checkpoint trained in target domain, evaluated in target domain after fine-tuning. Source and target domains are different boards in Pentalath.}
\label{Table:PentalathBoardsFinetuned}
\vspace{6pt}
\centering
\begin{tabular}{@{}lrr@{}}
\toprule
\textbf{Game: Pentalath} & \multicolumn{2}{c}{Target Domain} \\
\cmidrule(lr){2-3}
Source Domain & HexHexBoard & HalfHexHexBoard \\
\midrule
HexHexBoard & - & 72.67\% \\
HalfHexHexBoard & 52.00\% & - \\
\bottomrule
\end{tabular}
\end{table}

\begin{table}[H]
\caption{Win percentage of MCTS with final checkpoint from source domain against MCTS with final checkpoint trained in target domain, evaluated in target domain after fine-tuning. Source and target domains are different boards in Yavalath.}
\label{Table:YavalathBoardsFinetuned}
\vspace{6pt}
\centering
\begin{tabular}{@{}lrrrrrr@{}}
\toprule
\textbf{Game: Yavalath} & \multicolumn{6}{c}{Target Domain} \\
\cmidrule(lr){2-7}
Source Domain & $3$$\times$$3$ & $4$$\times$$4$ & $5$$\times$$5$ & $6$$\times$$6$ & $7$$\times$$7$ & $8$$\times$$8$ \\
\midrule
$3$$\times$$3$ & - & 39.50\% & 67.00\% & 57.33\% & 57.67\% & 70.67\% \\
$4$$\times$$4$ & 46.33\% & - & 73.00\% & 55.17\% & 60.67\% & 58.67\% \\
$5$$\times$$5$ & 49.00\% & 49.50\% & - & 52.00\% & 63.00\% & 58.33\% \\
$6$$\times$$6$ & 46.17\% & 54.83\% & 69.83\% & - & 65.00\% & 60.33\% \\
$7$$\times$$7$ & 52.00\% & 41.17\% & 68.83\% & 67.67\% & - & 74.00\% \\
$8$$\times$$8$ & 45.00\% & 67.50\% & 66.00\% & 53.17\% & 44.67\% & - \\
\bottomrule
\end{tabular}
\end{table}

\begin{table}[H]
\caption{Win percentage of MCTS with final checkpoint from source domain against MCTS with final checkpoint trained in target domain, evaluated in target domain after fine-tuning with reinitialised final layers. Source and target domains are different boards in Breakthrough.}
\label{Table:BreakthroughBoardsFinetunedReinit}
\vspace{6pt}
\centering
\begin{tabular}{@{}lrrrrrr@{}}
\toprule
\textbf{Game: Breakthrough} & \multicolumn{6}{c}{Target Domain} \\
\cmidrule(lr){2-7}
Source Domain & Square6 & Square8 & Square10 & Hexagon4 & Hexagon6 & Hexagon8 \\
\midrule
Square6 & - & 92.67\% & 96.33\% & 48.33\% & 66.67\% & 65.33\% \\
Square8 & 57.00\% & - & 88.33\% & 49.67\% & 60.33\% & 65.33\% \\
Square10 & 52.33\% & 53.33\% & - & 49.33\% & 42.33\% & 38.00\% \\
Hexagon4 & 47.67\% & 77.67\% & 95.00\% & - & 84.33\% & 73.00\% \\
Hexagon6 & 53.00\% & 86.67\% & 68.67\% & 49.67\% & - & 74.00\% \\
Hexagon8 & 52.33\% & 66.00\% & 93.33\% & 52.00\% & 74.00\% & - \\
\bottomrule
\end{tabular}
\end{table}

\begin{table}[H]
\caption{Win percentage of MCTS with final checkpoint from source domain against MCTS with final checkpoint trained in target domain, evaluated in target domain after fine-tuning with reinitialised final layers. Source and target domains are different boards in Broken Line.}
\label{Table:BrokenLineBoardsFinetunedReinit}
\vspace{6pt}
\centering
\begin{tabular}{@{}lrrrr@{}}
\toprule
\textbf{Game: Broken Line} & \multicolumn{4}{c}{Target Domain} \\
\cmidrule(lr){2-5}
Source Domain & LineSize3Hex & LineSize4Hex & LineSize5Square & LineSize6Square \\
\midrule
LineSize3Hex & - & 49.00\% & 50.00\% & 50.00\% \\
LineSize4Hex & 49.00\% & - & 50.00\% & 49.83\% \\
LineSize5Square & 50.00\% & 50.50\% & - & 50.00\% \\
LineSize6Square & 49.67\% & 49.67\% & 49.67\% & - \\
\bottomrule
\end{tabular}
\end{table}

\begin{table}[H]
\caption{Win percentage of MCTS with final checkpoint from source domain against MCTS with final checkpoint trained in target domain, evaluated in target domain after fine-tuning with reinitialised final layers. Source and target domains are different boards in Diagonal Hex.}
\label{Table:DiagonalHexBoardsFinetunedReinit}
\vspace{6pt}
\centering
\begin{tabular}{@{}lrrrrr@{}}
\toprule
\textbf{Game: Diagonal Hex} & \multicolumn{5}{c}{Target Domain} \\
\cmidrule(lr){2-6}
Source Domain & $7$$\times$$7$ & $9$$\times$$9$ & $11$$\times$$11$ & $13$$\times$$13$ & $19$$\times$$19$ \\
\midrule
$7$$\times$$7$ & - & 51.00\% & 86.67\% & 100.00\% & 100.00\% \\
$9$$\times$$9$ & 50.33\% & - & 89.67\% & 100.00\% & 100.00\% \\
$11$$\times$$11$ & 51.33\% & 46.83\% & - & 100.00\% & 100.00\% \\
$13$$\times$$13$ & 54.83\% & 49.67\% & 15.00\% & - & 53.00\% \\
$19$$\times$$19$ & 52.83\% & 49.33\% & 6.50\% & 45.17\% & - \\
\bottomrule
\end{tabular}
\end{table}

\begin{table}[H]
\caption{Win percentage of MCTS with final checkpoint from source domain against MCTS with final checkpoint trained in target domain, evaluated in target domain after fine-tuning with reinitialised final layers. Source and target domains are different boards in Gomoku.}
\label{Table:GomokuBoardsFinetunedReinit}
\vspace{6pt}
\centering
\begin{tabular}{@{}lrrrr@{}}
\toprule
\textbf{Game: Gomoku} & \multicolumn{4}{c}{Target Domain} \\
\cmidrule(lr){2-5}
Source Domain & $9$$\times$$9$ & $13$$\times$$13$ & $15$$\times$$15$ & $19$$\times$$19$ \\
\midrule
$9$$\times$$9$ & - & 68.00\% & 65.33\% & 70.00\% \\
$13$$\times$$13$ & 61.83\% & - & 65.33\% & 74.00\% \\
$15$$\times$$15$ & 59.33\% & 58.33\% & - & 71.67\% \\
$19$$\times$$19$ & 55.33\% & 55.17\% & 57.17\% & - \\
\bottomrule
\end{tabular}
\end{table}

\begin{table}[H]
\caption{Win percentage of MCTS with final checkpoint from source domain against MCTS with final checkpoint trained in target domain, evaluated in target domain after fine-tuning with reinitialised final layers. Source and target domains are different variants of Hex.}
\label{Table:HexBoardsFinetunedReinit}
\vspace{6pt}
\centering
\begin{tabular}{@{}lrrrrrr@{}}
\toprule
\textbf{Game: Hex} & \multicolumn{6}{c}{Target Domain} \\
\cmidrule(lr){2-7}
Source Domain & $7$$\times$$7$ & $9$$\times$$9$ & $11$$\times$$11$ & $13$$\times$$13$ & $19$$\times$$19$ & $11$$\times$$11$ Misere \\
\midrule
$7$$\times$$7$ & - & 68.00\% & 74.00\% & 100.00\% & 100.00\% & 87.33\% \\
$9$$\times$$9$ & 49.00\% & - & 72.67\% & 99.67\% & 100.00\% & 96.33\% \\
$11$$\times$$11$ & 49.00\% & 50.33\% & - & 100.00\% & 100.00\% & 93.00\% \\
$13$$\times$$13$ & 51.33\% & 41.33\% & 40.00\% & - & 100.00\% & 73.33\% \\
$19$$\times$$19$ & 49.67\% & 43.00\% & 36.00\% & 99.00\% & - & 0.67\% \\
$11$$\times$$11$ Misere & 47.00\% & 37.67\% & 41.00\% & 100.00\% & 84.00\% & - \\
\bottomrule
\end{tabular}
\end{table}

\begin{table}[H]
\caption{Win percentage of MCTS with final checkpoint from source domain against MCTS with final checkpoint trained in target domain, evaluated in target domain after fine-tuning with reinitialised final layers. Source and target domains are different boards in HeXentafl.}
\label{Table:HeXentaflBoardsFinetunedReinit}
\vspace{6pt}
\centering
\begin{tabular}{@{}lrr@{}}
\toprule
\textbf{Game: HeXentafl} & \multicolumn{2}{c}{Target Domain} \\
\cmidrule(lr){2-3}
Source Domain & $4$$\times$$4$ & $5$$\times$$5$ \\
\midrule
$4$$\times$$4$ & - & 52.17\% \\
$5$$\times$$5$ & 43.17\% & - \\
\bottomrule
\end{tabular}
\end{table}

\begin{table}[H]
\caption{Win percentage of MCTS with final checkpoint from source domain against MCTS with final checkpoint trained in target domain, evaluated in target domain after fine-tuning with reinitialised final layers. Source and target domains are different boards in Konane.}
\label{Table:KonaneBoardsFinetunedReinit}
\vspace{6pt}
\centering
\begin{tabular}{@{}lrrrr@{}}
\toprule
\textbf{Game: Konane} & \multicolumn{4}{c}{Target Domain} \\
\cmidrule(lr){2-5}
Source Domain & $6$$\times$$6$ & $8$$\times$$8$ & $10$$\times$$10$ & $12$$\times$$12$ \\
\midrule
$6$$\times$$6$ & - & 54.00\% & 76.33\% & 98.33\% \\
$8$$\times$$8$ & 51.67\% & - & 95.67\% & 99.33\% \\
$10$$\times$$10$ & 50.67\% & 36.00\% & - & 99.00\% \\
$12$$\times$$12$ & 51.33\% & 14.67\% & 38.00\% & - \\
\bottomrule
\end{tabular}
\end{table}

\begin{table}[H]
\caption{Win percentage of MCTS with final checkpoint from source domain against MCTS with final checkpoint trained in target domain, evaluated in target domain after fine-tuning with reinitialised final layers. Source and target domains are different boards in Pentalath.}
\label{Table:PentalathBoardsFinetunedReinit}
\vspace{6pt}
\centering
\begin{tabular}{@{}lrr@{}}
\toprule
\textbf{Game: Pentalath} & \multicolumn{2}{c}{Target Domain} \\
\cmidrule(lr){2-3}
Source Domain & HexHexBoard & HalfHexHexBoard \\
\midrule
HexHexBoard & - & 65.67\% \\
HalfHexHexBoard & 51.67\% & - \\
\bottomrule
\end{tabular}
\end{table}

\begin{table}[H]
\caption{Win percentage of MCTS with final checkpoint from source domain against MCTS with final checkpoint trained in target domain, evaluated in target domain after fine-tuning with reinitialised final layers. Source and target domains are different boards in Yavalath.}
\label{Table:YavalathBoardsFinetunedReinit}
\vspace{6pt}
\centering
\begin{tabular}{@{}lrrrrrr@{}}
\toprule
\textbf{Game: Yavalath} & \multicolumn{6}{c}{Target Domain} \\
\cmidrule(lr){2-7}
Source Domain & $3$$\times$$3$ & $4$$\times$$4$ & $5$$\times$$5$ & $6$$\times$$6$ & $7$$\times$$7$ & $8$$\times$$8$ \\
\midrule
$3$$\times$$3$ & - & 50.50\% & 56.17\% & 65.67\% & 72.00\% & 70.67\% \\
$4$$\times$$4$ & 48.33\% & - & 69.00\% & 68.50\% & 63.00\% & 47.33\% \\
$5$$\times$$5$ & 50.33\% & 51.67\% & - & 44.00\% & 57.67\% & 44.33\% \\
$6$$\times$$6$ & 53.17\% & 60.50\% & 68.17\% & - & 68.33\% & 53.67\% \\
$7$$\times$$7$ & 49.33\% & 54.83\% & 68.83\% & 56.67\% & - & 59.00\% \\
$8$$\times$$8$ & 51.17\% & 43.33\% & 57.67\% & 45.83\% & 68.00\% & - \\
\bottomrule
\end{tabular}
\end{table}

\section{Detailed Results -- Zero-shot Transfer Between Games}

Tables \ref{Table:line-completiongamesZeroShot}-\ref{Table:DiagonalHexCrossGameZeroShot} provide detailed results for zero-shot transfer evaluations, where source domains are different games from target domains (not just different variants).

\begin{table}[H]
\caption{Win percentage of MCTS with final checkpoint from source domain against MCTS with final checkpoint trained in target domain, evaluated in target domain (zero-shot transfer). Source and target domains are different line-completion games.}
\label{Table:line-completiongamesZeroShot}
\vspace{6pt}
\centering
\begin{tabular}{@{}lrrrrrr@{}}
\toprule
 & \multicolumn{6}{c}{Target Domain} \\
\cmidrule(lr){2-7}
Source Domain & Connect6 & Dai Hasami Shogi & Gomoku & Pentalath & Squava & Yavalath \\
\midrule
Connect6 & - & 0.00\% & 2.33\% & 0.00\% & 1.00\% & 0.33\% \\
Dai Hasami Shogi & 0.67\% & - & 1.33\% & 0.00\% & 0.67\% & 1.67\% \\
Gomoku & 36.67\% & 0.00\% & - & 0.33\% & 2.67\% & 1.33\% \\
Pentalath & 11.67\% & 0.00\% & 4.33\% & - & 2.00\% & 1.33\% \\
Squava & 16.00\% & 0.00\% & 0.33\% & 0.00\% & - & 2.00\% \\
Yavalath & 0.00\% & 0.00\% & 0.00\% & 0.33\% & 1.67\% & - \\
\bottomrule
\end{tabular}
\end{table}

\begin{table}[H]
\caption{Win percentage of MCTS with final checkpoint from source domain against MCTS with final checkpoint trained in target domain, evaluated in target domain (zero-shot transfer). Source and target domains are different Shogi variants.}
\label{Table:ShogivariantsZeroShot}
\vspace{6pt}
\centering
\begin{tabular}{@{}lrrrr@{}}
\toprule
 & \multicolumn{4}{c}{Target Domain} \\
\cmidrule(lr){2-5}
Source Domain & Hasami Shogi & Kyoto Shogi & Minishogi & Shogi \\
\midrule
Hasami Shogi & - & 1.33\% & 0.33\% & 52.67\% \\
Kyoto Shogi & 39.83\% & - & 3.00\% & 44.67\% \\
Minishogi & 47.17\% & 16.17\% & - & 97.00\% \\
Shogi & 23.83\% & 1.67\% & 0.00\% & - \\
\bottomrule
\end{tabular}
\end{table}

\begin{table}[H]
\caption{Win percentage of MCTS with final checkpoint from \textit{Broken Line} variants against MCTS with final checkpoint trained in target domain, evaluated in target domain (zero-shot transfer). Target domains are different line completion games.}
\label{Table:BrokenLineCrossGameZeroShot}
\vspace{6pt}
\centering
\begin{tabular}{@{}lrrrrrr@{}}
\toprule
 & \multicolumn{6}{c}{Target Domain} \\
\cmidrule(lr){2-7}
Source (Broken Line) & Connect6 & Dai Hasami Shogi & Gomoku & Pentalath & Squava & Yavalath \\
\midrule
LineSize3Hex & 0.00\% & 0.00\% & 0.00\% & 0.00\% & 0.67\% & 1.67\% \\
LineSize4Hex & 0.00\% & 0.00\% & 0.00\% & 0.00\% & 0.33\% & 1.67\% \\
LineSize5Square & 31.33\% & 0.00\% & 1.00\% & 0.33\% & 0.67\% & 1.33\% \\
LineSize6Square & 32.00\% & 0.00\% & 1.00\% & 1.67\% & 0.33\% & 2.00\% \\
\bottomrule
\end{tabular}
\end{table}

\begin{table}[H]
\caption{Win percentage of MCTS with final checkpoint from \textit{Diagonal Hex} variants against MCTS with final checkpoint trained in target domain, evaluated in target domain (zero-shot transfer). Target domains are different variants of \textit{Hex}.}
\label{Table:DiagonalHexCrossGameZeroShot}
\vspace{6pt}
\centering
\begin{tabular}{@{}lrrrrrr@{}}
\toprule
 & \multicolumn{6}{c}{Target (Hex)} \\
\cmidrule(lr){2-7}
Source (Diagonal Hex) & $7$$\times$$7$ & $9$$\times$$9$ & $11$$\times$$11$ & $11$$\times$$11$ Misere & $13$$\times$$13$ & $19$$\times$$19$ \\
\midrule
$7$$\times$$7$ & 0.00\% & 0.00\% & 0.00\% & 0.00\% & 0.00\% & 10.33\% \\
$9$$\times$$9$ & 0.00\% & 0.00\% & 0.00\% & 0.00\% & 0.00\% & 15.67\% \\
$11$$\times$$11$ & 0.00\% & 0.00\% & 0.00\% & 0.00\% & 0.00\% & 28.33\% \\
$13$$\times$$13$ & 0.00\% & 0.00\% & 0.00\% & 0.00\% & 0.00\% & 9.00\% \\
$19$$\times$$19$ & 0.00\% & 0.00\% & 0.00\% & 0.00\% & 0.00\% & 24.33\% \\
\bottomrule
\end{tabular}
\end{table}

\section{Detailed Results -- Transfer Between Games With Fine-tuning}

Tables \ref{Table:line-completiongamesAutotune}-\ref{Table:DiagonalHexCrossGameAutotune} provide detailed results for evaluations of transfer performance after fine-tuning, where source domains are different games from target domains (not just different variants). Note that in these cases, the two models that play against each other do not always have exactly the same number of trainable parameters. For hidden convolutional layers, we always use twice as many channels as the number of channels in a game's state tensor representation, and this is not modified when transferring to a new domain. This means that if a source domain has a greater number of channels in its state tensor representation than the target domain, the transferred model will also still use more channels in its hidden convolutional layers than the baseline model, and vice versa when the source domain has fewer state channels. Tables \ref{Table:line-completiongamesNoAutotune}-\ref{Table:DiagonalHexCrossGameNoAutotune} provide additional results where we adjust the number of channels of hidden convolutional layers when transferring models, prior to fine-tuning, for a more ``fair'' evaluation in terms of network size.

\begin{table}[H]
\caption{Win percentage of MCTS with final checkpoint from source domain against MCTS with final checkpoint trained in target domain, evaluated in target domain (after fine-tuning). Source and target domains are different line-completion games.}
\label{Table:line-completiongamesAutotune}
\vspace{6pt}
\centering
\begin{tabular}{@{}lrrrrrr@{}}
\toprule
 & \multicolumn{6}{c}{Target Domain} \\
\cmidrule(lr){2-7}
Source Domain & Connect6 & Dai Hasami Shogi & Gomoku & Pentalath & Squava & Yavalath \\
\midrule
Connect6 & - & 54.00\% & 53.17\% & 58.33\% & 45.50\% & 63.33\% \\
Dai Hasami Shogi & 54.67\% & - & 54.50\% & 53.67\% & 48.00\% & 72.33\% \\
Gomoku & 95.00\% & 50.17\% & - & 59.33\% & 48.50\% & 49.67\% \\
Pentalath & 92.00\% & 53.33\% & 57.67\% & - & 46.67\% & 50.50\% \\
Squava & 94.33\% & 51.00\% & 56.67\% & 64.00\% & - & 75.17\% \\
Yavalath & 43.00\% & 53.33\% & 56.00\% & 56.00\% & 45.00\% & - \\
\bottomrule
\end{tabular}
\end{table}

\begin{table}[H]
\caption{Win percentage of MCTS with final checkpoint from source domain against MCTS with final checkpoint trained in target domain, evaluated in target domain (after fine-tuning). Source and target domains are different Shogi variants.}
\label{Table:ShogivariantsAutotune}
\vspace{6pt}
\centering
\begin{tabular}{@{}lrrrr@{}}
\toprule
 & \multicolumn{4}{c}{Target Domain} \\
\cmidrule(lr){2-5}
Source Domain & Hasami Shogi & Kyoto Shogi & Minishogi & Shogi \\
\midrule
Hasami Shogi & - & 38.17\% & 40.17\% & 89.00\% \\
Kyoto Shogi & 45.67\% & - & 35.67\% & 70.00\% \\
Minishogi & 52.00\% & 63.17\% & - & 86.67\% \\
Shogi & 49.67 & 75.83\% & 36.00\% & - \\
\bottomrule
\end{tabular}
\end{table}

\begin{table}[H]
\caption{Win percentage of MCTS with final checkpoint from \textit{Broken Line} variants against MCTS with final checkpoint trained in target domain, evaluated in target domain (after fine-tuning). Target domains are different line completion games.}
\label{Table:BrokenLineCrossGameAutotune}
\vspace{6pt}
\centering
\begin{tabular}{@{}lrrrrrr@{}}
\toprule
 & \multicolumn{6}{c}{Target Domain} \\
\cmidrule(lr){2-7}
Source (Broken Line) & Connect6 & Dai Hasami Shogi & Gomoku & Pentalath & Squava & Yavalath \\
\midrule
LineSize3Hex & 56.00\% & 52.00\% & 54.83\% & 53.67\% & 49.00\% & 47.17\% \\
LineSize4Hex & 64.67\% & 52.83\% & 53.83\% & 56.33\% & 48.00\% & 68.00\% \\
LineSize5Square & 88.67\% & 53.67\% & 53.83\% & 66.00\% & 47.33\% & 64.00\% \\
LineSize6Square & 90.00\% & 52.33\% & 50.67\% & 52.00\% & 46.00\% & 58.67\% \\
\bottomrule
\end{tabular}
\end{table}

\begin{table}[H]
\caption{Win percentage of MCTS with final checkpoint from \textit{Diagonal Hex} variants against MCTS with final checkpoint trained in target domain, evaluated in target domain (after fine-tuning). Target domains are different variants of Hex.}
\label{Table:DiagonalHexCrossGameAutotune}
\vspace{6pt}
\centering
\begin{tabular}{@{}lrrrrrr@{}}
\toprule
 & \multicolumn{6}{c}{Target (Hex)} \\
\cmidrule(lr){2-7}
Source (Diagonal Hex) & $7$$\times$$7$ & $9$$\times$$9$ & $11$$\times$$11$ & $11$$\times$$11$ Misere & $13$$\times$$13$ & $19$$\times$$19$ \\
\midrule
$7$$\times$$7$ & 51.00\% & 59.00\% & 15.33\% & 48.67\% & 99.33\% & 80.00\% \\
$9$$\times$$9$ & 51.67\% & 50.33\% & 19.00\% & 53.33\% & 100.00\% & 70.67\% \\
$11$$\times$$11$ & 46.00\% & 19.67\% & 6.33\% & 23.00\% & 97.33\% & 40.67\% \\
$13$$\times$$13$ & 47.00\% & 56.67\% & 5.33\% & 0.67\% & 0.00\% & 20.00\% \\
$19$$\times$$19$ & 45.00\% & 54.67\% & 28.67\% & 1.67\% & 0.00\% & 45.33\% \\
\bottomrule
\end{tabular}
\end{table}

\begin{table}[H]
\caption{Win percentage of MCTS with final checkpoint from source domain against MCTS with final checkpoint trained in target domain, evaluated in target domain (after fine-tuning). Source and target domains are different line-completion games.}
\label{Table:line-completiongamesNoAutotune}
\vspace{6pt}
\centering
\begin{tabular}{@{}lrrrrrr@{}}
\toprule
 & \multicolumn{6}{c}{Target Domain} \\
\cmidrule(lr){2-7}
Source Domain & Connect6 & Dai Hasami Shogi & Gomoku & Pentalath & Squava & Yavalath \\
\midrule
Connect6 & - & 53.83\% & 57.50\% & 69.00\% & 51.00\% & 74.00\% \\
Dai Hasami Shogi & 55.33\% & - & 57.67\% & 59.00\% & 48.67\% & 68.83\% \\
Gomoku & 93.00\% & 52.00\% & - & 60.33\% & 46.00\% & 62.67\% \\
Pentalath & 76.67\% & 48.33\% & 59.83\% & - & 47.33\% & 58.33\% \\
Squava & 40.67\% & 50.00\% & 58.17\% & 58.67\% & - & 69.50\% \\
Yavalath & 70.33\% & 52.00\% & 51.67\% & 53.00\% & 49.00\% & - \\
\bottomrule
\end{tabular}
\end{table}

\begin{table}[H]
\caption{Win percentage of MCTS with final checkpoint from source domain against MCTS with final checkpoint trained in target domain, evaluated in target domain (after fine-tuning, with number of channels in hidden convolutional layers adjusted to be equal). Source and target domains are different Shogi variants.}
\label{Table:ShogivariantsNoAutotune}
\vspace{6pt}
\centering
\begin{tabular}{@{}lrrrr@{}}
\toprule
 & \multicolumn{4}{c}{Target Domain} \\
\cmidrule(lr){2-5}
Source Domain & Hasami Shogi & Kyoto Shogi & Minishogi & Shogi \\
\midrule
Hasami Shogi & - & 35.83\% & 34.67\% & 67.33\% \\
Kyoto Shogi & 48.00\% & - & 33.67\% & 63.33\% \\
Minishogi & 50.00\% & 58.00\% & - & 65.33\% \\
Shogi & 49.67\% & 45.67\% & 45.67\% & - \\
\bottomrule
\end{tabular}
\end{table}

\begin{table}[H]
\caption{Win percentage of MCTS with final checkpoint from \textit{Broken Line} variants against MCTS with final checkpoint trained in target domain, evaluated in target domain (after fine-tuning, with number of channels in hidden convolutional layers adjusted to be equal). Target domains are different line completion games.}
\label{Table:BrokenLineCrossGameNoAutotune}
\vspace{6pt}
\centering
\begin{tabular}{@{}lrrrrrr@{}}
\toprule
 & \multicolumn{6}{c}{Target Domain} \\
\cmidrule(lr){2-7}
Source (Broken Line) & Connect6 & Dai Hasami Shogi & Gomoku & Pentalath & Squava & Yavalath \\
\midrule
LineSize3Hex & 46.67\% & 50.00\% & 48.17\% & 56.00\% & 45.67\% & 74.17\% \\
LineSize4Hex & 45.67\% & 54.33\% & 52.17\% & 60.00\% & 48.33\% & 66.67\% \\
LineSize5Square & 94.00\% & 49.33\% & 54.33\% & 63.00\% & 48.33\% & 70.00\% \\
LineSize6Square & 82.67\% & 49.67\% & 50.17\% & 47.33\% & 47.00\% & 72.00\% \\
\bottomrule
\end{tabular}
\end{table}

\begin{table}[H]
\caption{Win percentage of MCTS with final checkpoint from \textit{Diagonal Hex} variants against MCTS with final checkpoint trained in target domain, evaluated in target domain (after fine-tuning, with number of channels in hidden convolutional layers adjusted to be equal). Target domains are different variants of Hex.}
\label{Table:DiagonalHexCrossGameNoAutotune}
\vspace{6pt}
\centering
\begin{tabular}{@{}lrrrrrr@{}}
\toprule
 & \multicolumn{6}{c}{Target (Hex)} \\
\cmidrule(lr){2-7}
Source (Diagonal Hex) & $7$$\times$$7$ & $9$$\times$$9$ & $11$$\times$$11$ & $11$$\times$$11$ Misere & $13$$\times$$13$ & $19$$\times$$19$ \\
\midrule
$7$$\times$$7$ & 50.67\% & 31.67\% & 56.33\% & 8.67\% & 99.67\% & 100.00\% \\
$9$$\times$$9$ & 47.00\% & 48.33\% & 42.00\% & 40.00\% & 98.67\% & 66.67\% \\
$11$$\times$$11$ & 47.67\% & 25.67\% & 42.33\% & 20.67\% & 87.00\% & 12.00\% \\
$13$$\times$$13$ & 49.33\% & 44.33\% & 8.33\% & 1.33\% & 0.00\% & 56.67\% \\
$19$$\times$$19$ & 50.00\% & 50.67\% & 28.33\% & 2.00\% & 0.00\% & 10.67\% \\
\bottomrule
\end{tabular}
\end{table}

\end{document}